\documentclass{IEEEtran}  

\IEEEoverridecommandlockouts                              
\usepackage[utf8]{inputenc}
\usepackage[T1]{fontenc}
\usepackage{graphicx}
\usepackage[namelimits]{amsmath} 
\usepackage{amssymb}             
\usepackage{amsfonts}            
\usepackage{mathrsfs}            
\usepackage{subfigure} 
\usepackage{booktabs}
\usepackage{cite}
\usepackage{url}
\usepackage{dsfont}
\usepackage{algorithm}
\usepackage{algpseudocode}
\usepackage{algorithmicx}
\usepackage{amsmath}
\usepackage[switch]{lineno}
\usepackage{bbding}

\usepackage{hyperref}

\usepackage{multirow}
\usepackage{textcomp,booktabs}
\usepackage[usenames,dvipsnames]{color}
\usepackage{colortbl}
\usepackage[dvipsnames]{xcolor}
\definecolor{mgray}{gray}{.9}
\definecolor{dgray}{gray}{.5}

\usepackage{algorithm}
\usepackage{algpseudocode}

\hypersetup{
    colorlinks=true,
    linkcolor=blue,
    filecolor=magenta,      
    urlcolor=cyan
}

\usepackage{soul,framed}

\usepackage{amsmath, amssymb}
\usepackage{amsthm}
\theoremstyle{plain}
\newtheorem{theorem}{Theorem}

\usepackage[colorinlistoftodos,bordercolor=blue,backgroundcolor=blue!20,linecolor=blue,textsize=scriptsize]{todonotes}
\title{Reinforced Refinement with Self-Aware Expansion for End-to-End Autonomous Driving}

\author{Haochen Liu, Tianyu Li, Haohan Yang,  Li Chen, Caojun Wang, Ke Guo, Haochen Tian, Hongchen Li,\\ Hongyang Li~\IEEEmembership{Senior Member, IEEE},  and Chen Lv$^{*}$,~\IEEEmembership{Senior Member, IEEE}
\thanks{This work was supported in part by  the Agency for Science, Technology and Research (A*STAR), Singapore, under the MTC Individual Research Grant (M22K2c0079), and the Ministry of Education (MOE), Singapore, under the Tier 2 Grant (MOE-T2EP50222-0002).}
\thanks{Haochen Liu, Haohan Yang, Ke Guo, and Chen Lv are with the School of Mechanical and Aerospace Engineering, Nanyang Technological University, Singapore, 639798.}
\thanks{Tianyu Li, Li Chen, Haochen Tian, and Hongyang Li are with the OpenDriveLab, the School of Computing and Data Science, The University of Hong Kong, Pokfulam, Hong Kong.}
\thanks{Tianyu Li, Caojun Wang and Hongchen Li are with the Shanghai Innovation Institute, Shanghai, 200231.}
\thanks{$^{*}$Corresponding author: Chen Lv, e-mail: lyuchen@ntu.edu.sg.}
}

\begin{document}
\maketitle

\begin{figure*}[t]
    \vspace{-0.5cm}
    \centering
    \includegraphics[width=\linewidth]{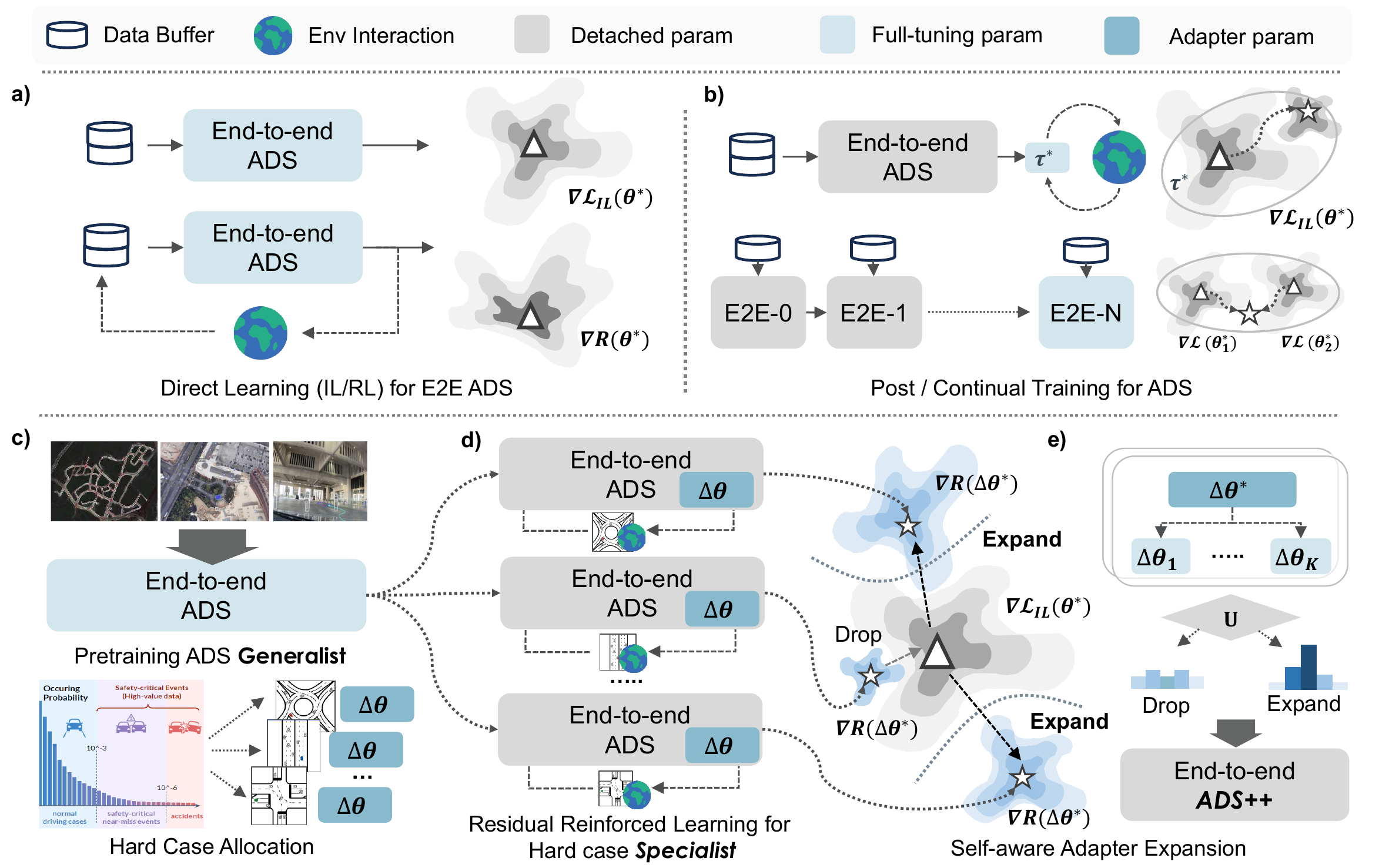}
    \caption[General Learning Framework in E2E ADS.]{\textbf{General Learning Framework in E2E ADS.} a) IL and RL pipelines directly train a generalist ADS from expert demonstrations or environmental rewards. b) Post-training approaches aim to improve E2E performance through online adaptation or continual learning in novel domains. c)–e) R2SE is established upon c) self-aware hard cases by pretrained ADS generalist. Then, d) a reinforced refinement paradigm in R2SE enhances performance on hard cases, while preserving generalist capabilities by e) adapter expansion. R2SE is applicable to model-agnostic E2E ADS. (Scenario figures refer to~\cite{li2024decode,dauner2024navsim})}
    \label{fig1}
    \vspace{-0.5cm}
\end{figure*}

\begin{abstract}
End-to-end autonomous driving has emerged as a promising paradigm for directly mapping sensor inputs to planning maneuvers using learning-based modular integrations. However, existing imitation learning (IL)-based models suffer from generalization to hard cases, and a lack of corrective feedback loop under post-deployment. While reinforcement learning (RL) offers a potential solution to tackle hard cases with optimality, it is often hindered by overfitting to specific driving cases, resulting in catastrophic forgetting of generalizable knowledge and sample inefficiency. To overcome these challenges, we propose Reinforced Refinement with Self-aware Expansion (R2SE), a novel learning pipeline that constantly refines hard domain while keeping generalizable driving policy for model-agnostic end-to-end driving systems. Through reinforcement fine-tuning and policy expansion that facilitates continuous improvement, R2SE features three key components: 1) Generalist Pretraining with hard-case allocation trains a generalist imitation learning (IL) driving system while dynamically identifying failure-prone cases for targeted refinement; 2) Residual Reinforced Specialist Fine-tuning optimizes residual corrections using reinforcement learning (RL) to improve performance in hard case domain while preserving global driving knowledge; 3) Self-aware Adapter Expansion dynamically integrates specialist policies back into the generalist model, enhancing continuous performance improvement. Experimental results in closed-loop simulation and real-world datasets demonstrate improvements in generalization, safety, and long-horizon policy robustness over state-of-the-art E2E systems, highlighting the effectiveness of reinforce refinement for scalable autonomous driving.
\end{abstract}


\section{Introduction}
\label{sec1}

\IEEEPARstart{A}{utonomous} driving has evolved into a pivotal advancement in mobility and transportation safety~\cite{chen2022milestones}, with end-to-end (E2E) autonomous driving systems (ADS) gaining increasing attention~\cite{chen2023end}. Due to its ability to unify perception, prediction, and planning into integrated modular learning-based system, E2E system could leverage direct sensory mapping with planning decisions. This significantly mitigates information loss and cascading errors that frequently occur in traditional modular pipelines.

E2E systems supervised by expert-annotated demonstrations further enhance human-like driving behaviors through imitation learning (IL)~\cite{gonzalez2015review,hu2023planning,jiang2023vad,liu2025hybrid}, which have witnessed wide deployment in scalable real-world scenarios. However, the paradigm of E2E autonomous driving systems remains constrained by generalization abilities in corner cases~\cite{sun2021corner}. These failure-prone corner cases can be categorized into two types, that is, rare samples unrepresented in the collected data~\cite{liu2024curse,feng2023dense}, and hard cases beyond the capacity of IL-pretrained E2E models~\cite{zhou2022dynamically}. Moreover, corner cases in IL-based E2E ADS can stem from various factors, such as perception failures (e.g., extreme weather, occlusions)~\cite{bolte2019towards} or challenging agent behavior in complex traffic interactions~\cite{huang2023gameformer}. While recent efforts have attempted to mitigate data rarity through world models~\cite{yang2024generalized,gao2024vista}, the hard cases characterized by under-fitted domain in E2E model capability remain underexplored. Addressing these challenges requires a targeted refinement of hard case, and extends the E2E model ability beyond IL distribution in hard driving scenarios.

Accordingly, reinforcement learning (RL) has gained a growing focus~\cite{aradi2020survey}. It enables long-horizon maneuvers beyond expert-level demonstrations by optimizing E2E driving policies through reward-driven feedback~\cite{li2024think2drive,liu2022improved,huang2022efficient},  making it a viable approach to addressing hard cases. However, success in solving specific hard cases~\cite{chen2019model} does not translate to generalizable scenarios, leading to significant overfitting compared to IL. Moreover, RL's reliance on massive real-world interactions~\cite{li2024think2drive} with high sample complexity~\cite{gao2025rad}, limiting practicality in ADS~\cite{chen2023end}. 

Beyond the learning paradigm, post-optimization and test-time training~\cite{huang2023dtpp,karkus2023diffstack,sima2025centaur,huang2022differentiable} have been explored to refine pre-trained E2E models in an online manner, providing a tentative solution to enhance test-time adaptability. However, these methods do not explicitly learn corner cases but rather fine-tune existing policies, often failing to ensure continuous improvement~\cite{cao2023continuous}. In contrast, continual learning (CL) is designed for streaming adaptation, allowing models to incorporate newly encountered cases over time~\cite{li2024decode,wu2022continual,yang2025human,cui2025sustainable}. Yet, CL-based methods, inherently upper-bounded by accumulated training data, are designed to boost adaptation towards new driving domains, rather than refine pre-existing hard cases and maintain generalization in collected data.

In response to these challenges, we propose \textbf{\underline{R}}einforced \textbf{\underline{R}}efinement with \textbf{\underline{S}}elf-aware \textbf{\underline{E}}xpansion (\textbf{R2SE}), a novel refinement learning paradigm for end-to-end (E2E) autonomous driving systems. R2SE pioneers the enhancement of hard-case domains while preserving generalized pretraining driving knowledge. Specifically, R2SE builds upon a model-agnostic, pretrained E2E autonomous driving generalist, dynamically identifying and leveraging failure cases during training. This allows R2SE to pinpoint failure-prone modules within the E2E system and execute targeted refinements for hard cases. To further optimize error corrections for hard-case specialists, we introduce a residual reinforced finetuning paradigm, while incorporating a set of low-ranked adapters~\cite{hu2022lora} to ensure both performance improvement in hard cases and retention of generalist knowledge. The self-aware adapter expansion mechanism adaptively integrates specialist policies back into the generalist model, leveraging uncertainty estimation to facilitate generalist policy improvement.
The main contributions are summarized as follows:

\begin{enumerate}
\item We propose R2SE, a model-agnostic reinforced refinement framework that improves scaled hard case performance for E2E autonomous driving system.

\item We introduce a self-aware refinement pipeline for hard case identification and improvement. Upon IL generalist, R2SE presents a residual reinforced adapter finetuning for hard case specialist. Adapter expansion further facilitates generalist improvement from reinforced specialists. 

\item R2SE is tested on multiple large-scale real-world benchmarks, and extensive testing results demonstrate its strong refinements in terms of generalization, safety, and long-term performance upon E2E systems.
\end{enumerate}

The remainder of this paper is organized as follows: Sec. \ref{sec2} reviews reinforced pipelines and E2E ADS. Sec. \ref{sec3} formulates the methodology of R2SE. Sec. \ref{sec4} presents comprehensive benchmark results. Finally, Sec. \ref{sec5} concludes the paper.

\section{Related Work}
\label{sec2}

\subsection{End-to-end Autonomous Driving System}
End-to-end pipelines follow a direct modular flow from raw sensor data to prediction and planning, guided by perception understanding \cite{chen2023end}. Emerged from one-model mapping, conventional E2E systems learn from sequential modular integrations. Representation misalignment challenge between modules in sequential E2E system facilitates the rise of BEV-based E2E perception \cite{li2022bevformer, zhang2022beverse, akan2022stretchbev}, which allows the modular integration within a unified BEV geometric space \cite{li2023delving}. Aligned representations enabled query-based modular integration toward planning-oriented objectives, leveraging either dense \cite{hu2023planning, ye2023fusionad, jia2023driveadapter} or sparse \cite{jiang2023vad} representations, to structure spatial-temporal information for improved trajectory reasoning. Confronted by sophisticated social interactions, interaction modeling between prediction and planning is considered in E2E framework by conditional prediction~\cite{chen2023deepemplanner} or game-theoretic reasoning~\cite{liu2025hybrid}. To formulate human-like planning distribution, a series of recent E2E pipelines turn to probabilistic planning leveraging categorical matching~\cite{chen2024vadv2} or diffusion process~\cite{liao2024diffusiondrive}. However, with primary focus on modular design to improve E2E alignment, E2E ADS still struggles to refine in corner cases from sampling-scarce rare scenarios or failure-prone hard cases. 

In response to the rarity challenge, the integration of vision language model (VLM) and world model (WM) has emerged as a novel extension in E2E ADS beyond modular design. VLMs have been explored as motion planners \cite{mao2023gpt, tian2024drivevlm}, routing agents \cite{mao2023language, sima2023drivelm}, offering interpretability and high-level reasoning capabilities. Similarly, world models aim to supplement data rarity by simulating diverse environments~\cite{gao2024vista}, evolving additional training samples~\cite{ma2024unleashing}, or guiding test-time optimizations~\cite{wang2024driving}. While these approaches enrich E2E systems with commonsense knowledge or generated scenarios to address data rarity, they do not explicitly tackle hard-case failures that exceed the capacity of behavioral cloning pipelines. In contrast, R2SE addresses the core challenges of E2E systems in handling hard cases by introducing a self-aware refinement paradigm. It dynamically identifies failure-prone scenarios and reasons about the specific modules. To ensure scalability, R2SE is designed to be model-agnostic, capable of operating on generalist E2E architectures without relying on specific modular designs, and shows strong potential for integration with VLMs or WMs that focus on complementary objectives.

\subsection{Reinforcement learning in E2E ADS}
Reinforcement learning (RL) has long been explored in ADS to optimize driving under suboptimal or unlabeled scenarios~\cite{kiran2021deep}, garnering increased attention upon E2E settings. Early research attempted to use RL to converge toward optimal E2E policies on a per-scenario basis~\cite{chen2019model}. However, such scenario-specific RL approaches often suffer from slow convergence. To improve efficiency, methods introduced human-in-the-loop corrections~\cite{wu2023human}, IL constraints~\cite{huang2022efficient}, demonstration buffers~\cite{liu2022improved} or pretraining~\cite{liu2024augmenting}. Still, RL-based models tend to overfit and struggle to generalize across diverse, real-world driving scenarios. Another line of RL research frames the RL agent as a driving expert within scaled simulation environments~\cite{chen2021learning}, particularly where expert demonstrations are unavailable~\cite{zhang2024feedback,zhang2021end}.  Faced with the challenge of slow convergence in complex scenarios, model-based RL leveraging latent world models~\cite{jia2023think} has shown promise in accelerating learning and improving simulated driving performance~\cite{wang2025adawm}. However, such approaches often suffer from a significant sim-to-real gap and exhibit behavioral mismatches from human driving, increasing difficulty to deploy in real-world E2E cases. 

To improve end-to-end performance in hard cases, recent pipelines have turned to RL by refinement after expert knowledge pretraining. Either through adapter-based learning from teacher models~\cite{jia2023driveadapter}, or fine-tuning within reconstructed 3D scenes~\cite{gao2025rad}, RL reports improved closed-loop performance for E2E system. However, these approaches often rely on identical architectures for adapter learning~\cite{liu2023tail}, or require costly reconstructed environments for online interaction. Additionally, they are prone to catastrophic forgetting of pretrained IL knowledge and frequently require replay strategies to retain prior capabilities~\cite{binici2022preventing}.
Contrarily, R2SE decouples hard cases by identifying failure-prone modules. We also argue that applying offline RL with safe constraints could receive scaled improvement, preserving IL generalization. A residual reinforcement fine-tuning mechanism allows specialist adapters to correct errors efficiently, enabling fast adaptation without costly online interactions and preventing knowledge forgetting.

\begin{figure*}[t]
    \centering
    \includegraphics[width=\linewidth]{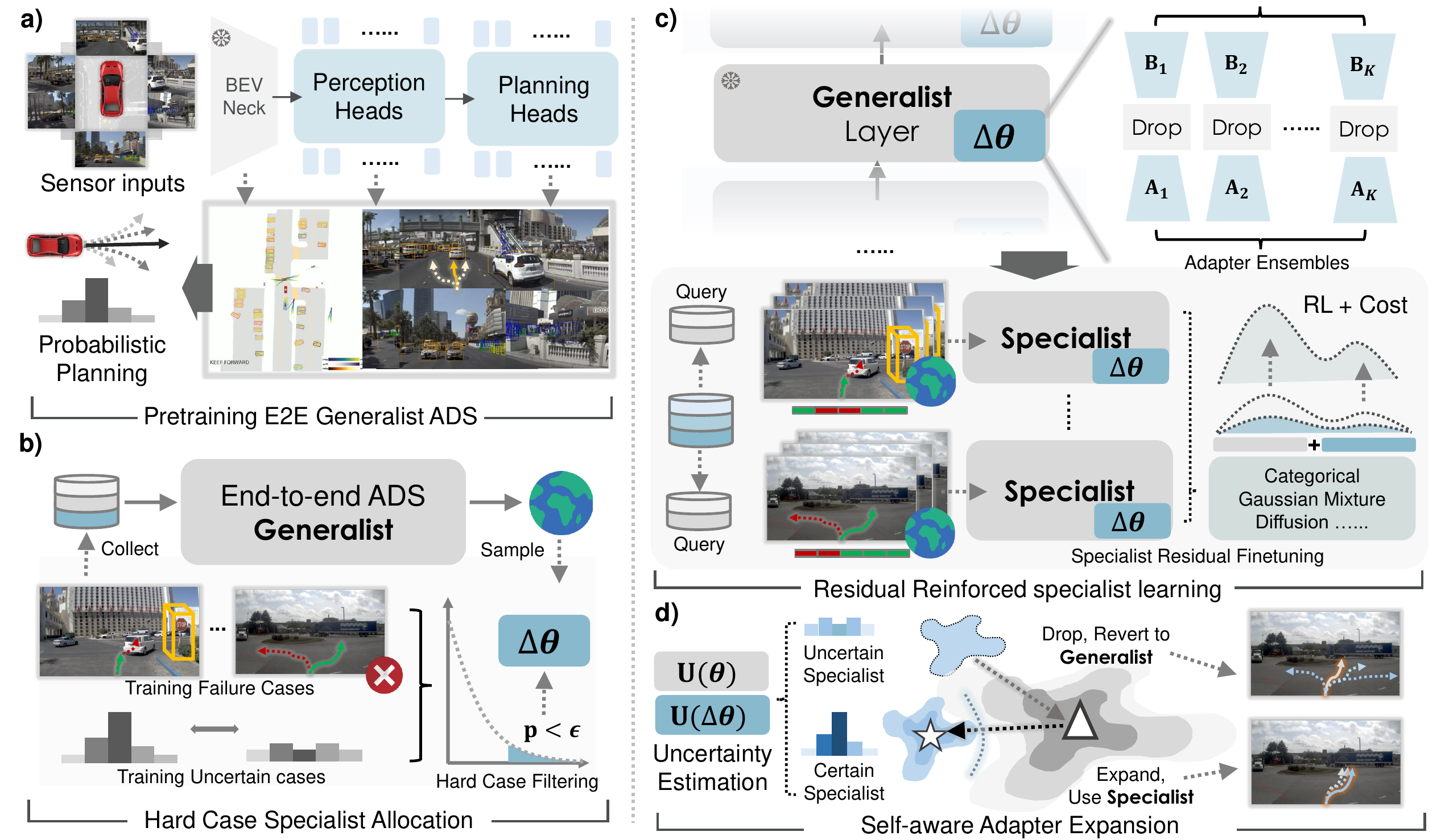}
    \vspace{-0.5cm}
    \caption[System overview of R2SE.]{\textbf{System overview of R2SE.} R2SE is a reinforced refinement framework designed to improve model-agnostic E2E autonomous driving through adapter expansion. a) It begins by pretraining a generalist policy $\pi_\theta$ using sensor inputs $\mathbf{X}$ and multi-head outputs $\mathbf{Y}$ for perception and planning. b) Hard cases are then identified based on a difficulty metric $\mathcal{F}_x$, capturing uncertain or failed behaviors, and sampled for specialization. c) Retrieved clips $\mathcal{D}_{\text{RL}}$ are used to finetune specialist adapters via residual reinforcement learning optimized using GRPO over process reward and cost. d) At test time, R2SE uses ensemble uncertainty with GPD-based density inference to adaptively expand between  specialist adapters and generalist policies.}
    \vspace{-0.5cm}
    \label{fig2}
\end{figure*}

\subsection{Self-aware learning and adaptation in ADS}
Self-awareness has emerged as a key topic, that enables model to generate feedback~\cite{wu2024self} and actively reinforce their performance~\cite{xie2023active}. In active learning, models adaptively select important samples during training~\cite{lu2024activead}, but such approaches are less explored in autonomous driving where full training data is typically available. Self-awareness has also been explored through game-theoretic self-play for state-based exploration~\cite{cusumano2025robust,zhang2024learning}, yet gaps remain in real-world driving scenarios to quantify the quality of feedback from the model’s own policy. Explicit feedback typically involves reward or cost functions during interactions~\cite{li2025finetuning}, while implicit feedback, such as uncertainty estimation~\cite{cao2021confidence} or entropy~\cite{janson2017monte}, has seen few exploration. Some recent works leverage these ideas in test-time training to minimize self-supervised costs~\cite{sima2025centaur}. Techniques like evidential learning~\cite{amini2020deep} and conformal prediction~\cite{straitouri2023improving} offer more accurate uncertainty estimation, but require specific output modeling. Most self-awareness frameworks rely on adaptation or fine-tuning, which risks catastrophic forgetting of pretrained knowledge. Continual adaptation attempts to mitigate this using replay buffers~\cite{wu2022continual}, gradient constraints~\cite{yang2025human}, or dynamic networks~\cite{li2024decode,cui2025sustainable}. While demand scenario-specific design and classification, they often fall short in scale in real-world driving logs. Despite recent advances in using efficient finetuning adapters~\cite{wang2024forecast} to handle style transfer~\cite{kothari2023motion} and out-of-distribution behaviors~\cite{diehl2024lord}, there remains a lack of adaptive pipelines for discovering and refining hard cases while mitigating forgetting.

R2SE leverages both explicit (cost-constrained) and implicit (inference-based) feedback to identify failure-prone cases and guide offline reinforced refinement of specialist modules. Implicit feedback further supports the self-aware adapter expansion to maintain generalist performance in R2SE, when specialists are uncertain. To enable efficient uncertainty estimation, R2SE offers a set of low-rank adapters ensembles without altering the original model output.

\section{Reinforced Refinement with Self-aware Expansion}
\label{sec3}
Fig. \ref{fig1} presents the general structure of our proposed R2SE framework, as each sub-module is detailed in Fig.  \ref{fig2}. R2SE leverages the reinforced pretraining-finetuning paradigm targeted on hard cases refinement for model-agnostic E2E ADS. The following Section \ref{sub1} will elaborate on the strategy for pretraining and hard case discovery (Fig.~\ref{fig1}c) on typical E2E ADS framework. R2SE further specializes hard domain capability by residual reinforced learning (Fig.~\ref{fig1}d) discussed in Section \ref{sub2}. Finally, Section \ref{sub3} outlines the self-aware adapter expansion integrated reinforced specialist back into improved E2E generalist (Fig.~\ref{fig1}e).

\subsection{Generalist Pretraining and Specialist Allocation}
\label{sub1}
As shown in Fig. \ref{fig2}, R2SE addresses the challenge of hard-case refinement in E2E autonomous driving systems by first establishing a generalist pretraining paradigm for model-agnostic E2E systems, followed by a quantitative process to identify and allocate failure-prone cases for specialists.

\subsubsection{Generalist Pretraining}
During the pretraining stage of R2SE, the model-agnostic E2E generalist would be built supposing a minimalist and deployable modular architecture for ADS, as shown in Fig.~\ref{fig2}a. The system receives streamed sensory inputs $\mathbf{X}$ over \(T_h\) steps, including panoptic-view images \(\mathbf{X}_{\text{Img}}\), ego state estimates \(\mathbf{X}_{\text{ego}}\), and LiDAR point clouds \(\mathbf{X}_{\text{LiDAR}}\). A BEV-based sensory neck abstracts and fuses multi-modal inputs \(\mathbf{X}\subseteq\{\mathbf{X}_{\text{Img}},\mathbf{X}_{\text{ego}},\mathbf{X}_{\text{LiDAR}}\}\) into an ego-centered geometric space \(H \times W\). BEV perception heads extract task-relevant features from BEV neck \(\mathbf{H}_{\text{BEV}}\), tracked agent \(\mathbf{H}_{\text{Agent}}\) for \(N_A\) agents, and online mapping \(\mathbf{H}_{\text{Map}}\) for \(N_M\) segments. These features are first pretrained on perception tasks for \(\hat{\mathbf{X}}_{\text{Agent}}\) and \(\hat{\mathbf{X}}_{\text{Map}}\) for subsequent planning.

In the second stage, queried perception features $\mathbf{H}$ are leveraged to inform end-to-end planning objectives. Confronted with diverse driving styles and behaviors from human demonstrations, R2SE configures a probabilistic planning head using a DETR-like model that queries perception features \(\mathbf{H}\subseteq\{\mathbf{H}_{\text{BEV}},\mathbf{H}_{\text{Agent}},\mathbf{H}_{\text{Map}}\}\). Given a set of \(M\) planning actions \(\mathbf{Y} = \{\mathbf{y}_1, \dots, \mathbf{y}_M\} \sim \pi_\theta(\mathbf{X}),\mathbf{y}=(x,y,\psi)_{1:T}\) over \(T\) steps, R2SE formulates the behavior cloning objective as distribution matching problem:
$\max\operatorname{KL}(\pi_{\text{exp}} \| \pi_{\theta})=\max\sum_{M}\pi_{\text{exp}}(\mathbf{X})\log(\pi_{\text{exp}}(\mathbf{X})/\pi_{\theta}(\mathbf{X}))$.
The E2E objective for pretraining the generalist in R2SE is therefore given by:
\begin{equation}
\label{equ:pretrain}
    \mathcal{L}_{\text{Pretrain}} = \mathcal{L}_{\text{Perception}} - \alpha\, \operatorname{KL}(\pi_{\text{exp}} \| \pi_{\theta})
\end{equation}
where the planning distribution \(\pi_{\theta}\) could be adapted to various probabilistic formulations for pretrained R2SE generalist.

\subsubsection{Hard Case Specialist Allocations}
Identifying failure-prone cases requires active exploration. Unlike active learning for annotation purposes, R2SE uncovers hard cases upon pretrained generalist. As in Fig. \ref{fig2}b, hard case allocation involves two key steps: (1) scoring case difficulty upon explicit and self-aware feedback $\mathcal{F}$ from the generalist, and (2) importance sampling for specialists. R2SE scores explicit feedback on log-replayed clips using perception and planning metrics to identify failures of the pretrained generalist $\pi_\theta$ on $\mathcal{D}_{\text{train}}$. With perception being more robust in E2E settings, it is loosely assessed via normalized loss: $\mathcal{F}_{\text{Per}}(\hat{\mathbf{X}}) = \mathcal{L}(\hat{\mathbf{X}}) / \max_{\hat{\mathbf{X}} \sim \mathcal{D}} (\hat{\mathbf{X}})$. While as the final output, planning feedback $\mathcal{F}_{\text{Plan}}(\hat{\mathbf{Y}},\mathbf{X})$ requires strict quantification using PDMScore (PDMS)~\cite{dauner2024navsim}, a comprehensive closed-loop metric with reference to state-of-the-art privileged planner~\cite{dauner2023parting}:
\begin{equation}
\label{equ:pdms}
\mathcal{F}_{\text{Plan}}(\hat{\mathbf{Y}},\mathbf{X}) = \prod_{m \in \{\text{NC}, \text{DAC}\}} \mathcal{F}^m\times\frac{
\sum_{w \in \{\text{EP}, \text{TTC}, \text{C}\}} \text{w}_w\mathcal{F}^w
}{
\sum_{w \in \{\text{EP}, \text{TTC}, \text{C}\}} \text{w}_w
}
\end{equation}
\begin{figure*}[t]
    \centering
    \includegraphics[width=\linewidth]{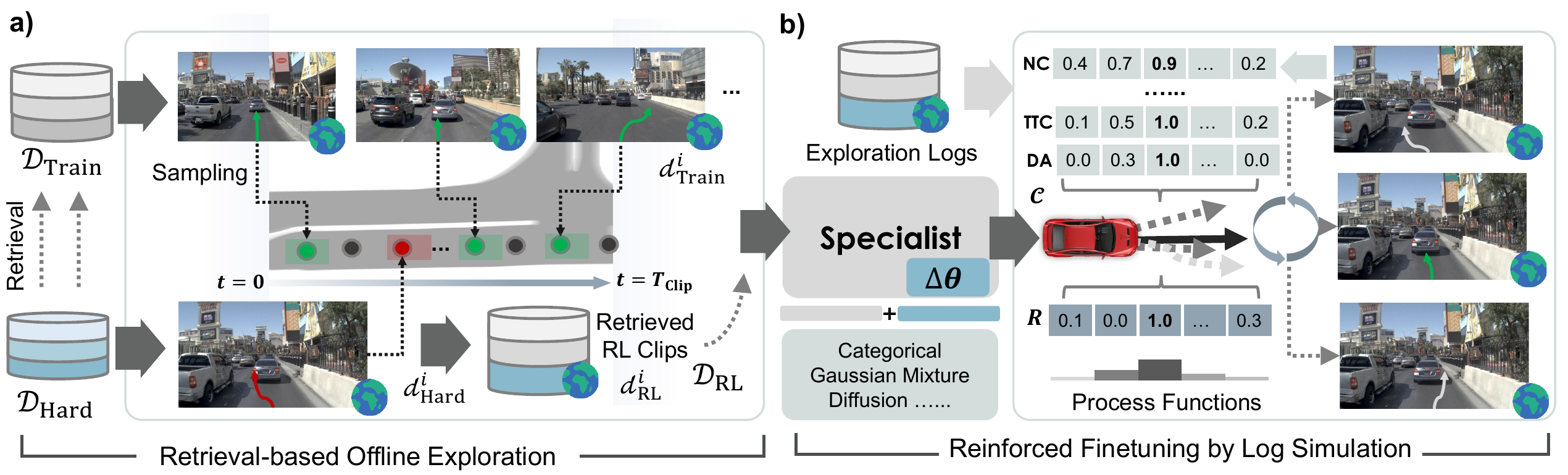}
    \vspace{-0.5cm}
    \caption[Residual Reinforced Specialist Learning in R2SE.]{\textbf{Residual Reinforced Specialist Learning in R2SE.} a) Retrieval-based exploration for $\mathcal{D}_{\text{RL}}$; b) GRPO-enabled reinforced finetuning. Set of process reward $\mathcal{R}$ and cost $\mathcal{C}$ functions evaluate each candidate planning policy $\hat{\mathbf{y}}\in\hat{\mathbf{Y}}$ from batched non-reactive simulations using exploration clips of $\mathcal{D}_{\text{RL}}$.}
    \vspace{-0.5cm}
    \label{fig4}
\end{figure*}
wherein a set of normalized (0–1) multipliers and weighted sub-metrics jointly assess planning safety, comfort, and progress from closed-loop logs. Another source of case difficulty in R2SE comes from self-aware feedback $\mathcal{F}_{\text{Self}}(\hat{\mathbf{Y}})$: E2E systems often exhibit fragile failures in scenarios where ADS exhibits increasing planning uncertainty~\cite{wang2024survey}, which can be quantified by SOTIF Shannon Entropy (CE)~\cite{peng2023sotif}:
\begin{equation}
    \mathcal{F}_{\text{Ent}}(\hat{\mathbf{Y}})=\sum_{m\in M}\pi_{\theta}(\hat{\mathbf{y}}_m|\mathbf{X})\cdot\log\pi_{\theta}(\hat{\mathbf{y}}_m|\mathbf{X})
\end{equation}

R2SE’s active quantification of hard cases, then follows a monotonic progression with increasing case difficulty:
\begin{equation}
\label{equ:case_score}
    \mathcal{F}_\textbf{X} = (1-\mathcal{F}_{\text{Plan}}(\hat{\mathbf{Y}},\mathbf{X})) + \beta_{\text{Per}}\mathcal{F}_{\text{Per}}(\hat{\mathbf{X}})+\beta_{\text{Ent}} \mathcal{F}_{\text{Ent}}(\hat{\mathbf{Y}})
\end{equation}
This further enables the active importance sampling process for hard case subset $\mathcal{D}_{\text{hard}}\subset\mathcal{D}_{\text{train}}$ ranking by:
\begin{equation}
\mathcal{D}_{\text{hard}} = \left\{ d_{\text{hard}}=(\mathbf{X},\mathbf{Y}) \in \mathcal{D}_{\text{train}} \mid \mathcal{F}_{\mathbf{X}} \leq \mathbb{Q}_{\epsilon}(\mathcal{F}) \right\}
\end{equation}
where $\mathbb{Q}_{\epsilon}$ denotes the $\epsilon^{\text{th}}$ percentile of case difficulty. The hard case allocation process of R2SE works in parallel with the generalist inference, and could be suitable to either offline testing or online interactions.

\subsection{Residual Reinforced Specialist Learning}
\label{sub2}
To fully optimize hard case specialists with subset $\mathcal{D}_{\text{hard}}$ while retaining generalist knowledge, R2SE takes the initiative in residual reinforced finetuning. As in Fig. \ref{fig2}c,  specialists are reinforced to optimize residual error supported by newly-deployed adapter ensembles.
\subsubsection{Specialist Adapter\label{adapter}} Efficient adaptation to hard case residuals can be challenging and variable. Hence, instead of relying on single adapter set, R2SE enhances specialist robustness through a mixture of $K$ adapter ensembles for residual learning. Expressly, R2SE specialist is refined by adapter ensembles $\Delta\theta=\{\Delta\theta_1,\dots,\Delta\theta_K\}$, where each subset adapter is defined by a set of low-ranked adapters (LoRAs). Consider certain specialist layer, the adapter is formulated by:
\begin{equation}
\mathbf{W}_{\text{spec}}=\texttt{detach}(\mathbf{W})+\frac{1}{r}\textbf{A}_k\textbf{B}_k
\end{equation}
where \( \mathbf{W} \in \mathbb{R}^{d \times d} \subset \theta \) denote the generalist weights, which are detached from gradient computations. The matrices \( \mathbf{A}_k \in \mathbb{R}^{d \times r} \) and \( \mathbf{B}_k \in \mathbb{R}^{r \times d} \), where \( \mathbf{A}_k, \mathbf{B}_k \subset \Delta\theta_k \) for each task \( k \in K \), represent the LoRA parameters. This formulation significantly reduces refinement computation to \( \mathcal{O}(2ndrK) \), in contrast to dynamic networks that incur \( \mathcal{O}(nd^2) \) cost, which scales with the number of tasks \( d \gg 2rK \).The learning process for the specialist adapters proceeds as:
\begin{equation}
    \Delta\theta \leftarrow \Delta\theta - \eta \cdot \frac{1}{K} \sum_{k \in K} \nabla(\Delta\theta_k) \mathcal{L}(\pi_{\theta + \Delta\theta_k}(\mathbf{X}))
\end{equation}
With learning rate $\eta$, each individual specialist adapter on top of the shared generalist model is updated independently across the hard-case tasks. This allows targeted adaptation without interfering with the shared generalist.

\subsubsection{Reinforced Finetuning} 
Specialist residuals on hard cases demand safe and efficient reinforcement learning (RL) exploration, while preserving generalist knowledge. R2SE addresses this challenge via a constrained Markov Decision Process (cMDP)~\cite{wachi2020safe} formulation with finite horizons. Specifically, we define the cMDP with transition tuples $(\mathcal{S}, \mathcal{A}, \mathcal{R}, \mathcal{C}, \gamma)$, where the input $\mathbf{X}=s \in \mathcal{S}$ and the predicted output $\hat{\mathbf{Y}}=a \in \mathcal{A}$. The RL objective of R2SE over a horizon of length $T$ is:
\begin{equation}
\label{equ:constraint}
    \max_{\pi} \; \mathbb{E}_{\pi} \left[ \sum_{t=0}^{T} \gamma^t \mathcal{R}(s_t, a_t) \right] 
    \quad \text{s.t.} \quad 
    \mathbb{E}_{\pi} \left[ \sum_{t=0}^{T} \gamma^t \mathcal{C}(s_t, a_t) \right] \leq \delta,
\end{equation}
where $\mathcal{R}$ and $\mathcal{C}$ represent the reward and cost functions specific to challenging cases. This constrained optimization problem can be relaxed into a value penalty~\cite{huang2022efficient} form:
\begin{equation}
\label{equ:lag}
\min_{\pi} -\sum_{t=0}^{T} \gamma^t \mathcal{R}(s_t, a_t) 
+ \lambda \left\| \sum_{t=0}^{T} \gamma^t \mathcal{C}(s_t, a_t) - \delta \right\|,
\end{equation}
where $\lambda$ is a penalty multiplier. Beyond solving this objective, R2SE further needs to optimize residuals over the entire trajectory $\hat{\mathbf{Y}}$, while preserving the pretrained generalist knowledge. To this end, R2SE refines upon GRPO~\cite{shao2024deepseekmath}  (Fig.~\ref{fig4}) to effectively solve Equ.~\ref{equ:lag}: 
\begin{equation}
\label{equ:grpo}
    \mathcal{L}_{\text{RL}}=\texttt{IS}(\mathbf{X})\sum_{t\in T}[-\mathcal{R}_t\log\pi_{\theta+\Delta\theta}(\hat{\mathbf{Y}}|\mathbf{X})+\lambda\mathcal{C}_t(\mathbf{X},\hat{\mathbf{Y}})]
\end{equation}
Following GRPO, we emit the clip term for simplicity, and $\text{KL}(\cdot||\cdot)$ term as it already presents in $\mathcal{L}_{\text{pretrain}}$. $\texttt{IS}(\mathbf{X})=
\pi_{\theta+\Delta\theta}(\hat{\mathbf{Y}}|\mathbf{X})/{\pi_{\theta}(\hat{\mathbf{Y}}|\mathbf{X})}$ terms for the importance sampling weight. Reward and cost terms follow the standardized process supervisions~\cite{shao2024deepseekmath} (Fig.~\ref{fig4}b) leveraging planning feedback: $\mathcal{R}_t=\mathbf{1}(t=T)\gamma^{T-t}\mathcal{F}_{\text{plan}}(\hat{\mathbf{Y}},\mathbf{X})$. The cost terms also follow a series of standardized process functions from sub-metrics: $\mathcal{C}_t=\mathbf{1}(t=T)\sum_{m+w}\gamma^{T-t}\mathcal{F}^{m+w}(\hat{\mathbf{Y}},\mathbf{X})$, where $m+v\in\{\text{NC}, \text{DAC},\text{EP}, \text{TTC}, \text{C}\}$ constraining R2SE's  safety, route compliance, comfort, and driving efficiency relatively. 

\begin{algorithm}[t]
\caption{R2SE Pretraining and Refinement\label{algo:r2se}}
\begin{algorithmic}[1]
\State \textbf{Input:} General dataset $\mathcal{D}_{\text{Train}}$; Generalist model $\pi_{\theta}$
\State \textcolor{gray}{ \texttt{1. Generalist Pretraining:}}
\State E2E pretraining: $\mathcal{L}_{\text{pretrain}}(\pi_{{\theta}}(\mathbf{X}), \mathbf{Y})$ by Equ.~\ref{equ:pretrain}

\State \textcolor{gray}{ \texttt{2. Hard Case Allocation:}}
\For{$\mathbf{X} \sim \mathcal{D}_{\text{Train}}/\texttt{Env}$}
    \State $\hat{\mathbf{Y}}=\pi_\theta(\mathbf{X})$  \textcolor{gray}{// Execute with recovery for Env cases}
    \State Inference case score: $\mathcal{F}_{\mathbf{X}}(\hat{\mathbf{Y}},\mathbf{X}) $ by Equ.~\ref{equ:case_score}
\EndFor
\State Importance sampling $\mathcal{D}_{\text{Hard}}\subset\mathcal{D}_{\text{Train}}$ with percentile $\epsilon$ 
\vspace{0.5em}

\State \textcolor{gray}{ \texttt{3. Specialist Adapter and Exploration:}}
\State Specialist Initialization $\pi_{\theta+\Delta\theta}$ by adapter ensembles $\Delta\theta$
\For{$d_{\text{Hard}} \in \mathcal{D}_{\text{Hard}}$}
    \State Retrieve corresponding full scene clip $d_{\text{scene}}$
    \State Sample $L$ additional clips $d^1_{\text{Train}}, ..., d^L_{\text{Train}} \sim \mathcal{D}_{\text{Train}}$
    \State Construct RL tuning set: $d_{\text{RL}} = d_{\text{scene}} \cup \bigcup_{n=1}^L d^n_{\text{Train}}$
    \State $\mathcal{D}_{\text{RL}}=\mathcal{D}_{\text{RL}}\cup d_{\text{RL}}$
\EndFor
\State \textcolor{gray}{ \texttt{4. Specialist Refinement:}}
\For{clip $d_{\text{RL}}\in\mathcal{D}_{\text{RL}}$}
    \State $\hat{\mathbf{Y}}=\pi_{\theta+\Delta\theta}(\mathbf{X})$; \textbf{for} $\hat{\mathbf{y}}\in\hat{\mathbf{Y}}$: Evaluate process reward $\mathcal{R}$ and costs $\mathcal{C}$ via non-reactive simulation by clip $d_{\text{RL}}$
    \State Optimize $\hat{\mathbf{Y}}$ with GRPO in Equ.~\ref{equ:grpo} and Equ.~\ref{equ:pretrain}: \State$\mathcal{L}_{\text{RL}}+\alpha_{\text{Pretrain}}\mathcal{L}_{\text{Pretrain}}$
\EndFor
\State Return refined specialist policies $\pi_{\theta+\Delta\theta}$ and $\mathcal{D}_{\text{Hard}}$
\end{algorithmic}
\end{algorithm}

The generality of R2SE toward model-agnostic policies is further supported by the adaptability of the GRPO reinforcement above, to various output formulations of $\log \pi_{\theta}$:

\begin{itemize}
    \item[(1)] For policies modeled as categorical or Gaussian mixture distributions~\cite{chen2024vadv2,li2024hydra}, i.e., $\pi_{\theta} \sim \texttt{Cat}(\mathbf{X})$ or $\texttt{GMM}(\mathbf{X})$, the RL objective reduces to the negative log-likelihood (NLL) of the target distribution.
    
    \item[(2)] For more recent approaches such as diffusion-based policies~\cite{liao2024diffusiondrive}, the RL objective is used to optimize the conditional guidance of $p(\hat{\mathbf{y}}^{T_{\text{diff}}})$, where each planning $\hat{\mathbf{y}} \in \hat{\mathbf{Y}}$ is modeled by conditional diffusion model (CDM):
\end{itemize}
\begin{equation}
\pi_{\theta}(\hat{\mathbf{y}}^0 \mid z) = \int p(\hat{\mathbf{y}}^{T_{\text{diff}}}) \prod_{i=1}^{T_{\text{diff}}} \pi_{\theta}(\hat{\mathbf{y}}^{i-1} \mid \hat{\mathbf{y}}^i, z) \, \mathrm{d}\hat{\mathbf{y}}^{1:T_{\text{diff}}} .
\end{equation}

Still, rolling-out RL in E2E sensory setups remains costly due to scene reconstruction and sensory simulation requirements. R2SE alleviates this challenge by reformulating rollout as an offline retrieval problem. As in Fig.~\ref{fig4}a, R2SE leverages the GRPO design with log simulation to provide dense process reward and cost signals in a single pass. Specifically, consider a hard case clip defined as $d_{\text{Hard}} = \{\mathbf{X}_t, \mathbf{Y}_t \mid t \in [-T_h, T]\} \subset \mathcal{D}_{\text{Hard}}$. R2SE directly retrieves full scenario clips associated with such hard cases. To support generalization, a subset of $L$ training samples $d_{\text{Train}} \sim \mathcal{D}_{\text{Train}}$ is also retrieved. The combined set forms the RL training set: $d_{\text{RL}} = \bigcup_{n=1}^N d^n_{\text{Train}} \cup d_{\text{Hard}}, \; d_{\text{RL}} \in \mathcal{D}_{\text{RL}}$. Similar to rehearsal-based methods~\cite{verwimp2021rehearsal}, this retrieval-based exploration efficiently shapes R2SE’s decision boundary using only logged data clips. Furthermore, GRPO enables dense optimization over all candidate policies through non-reactive simulations~\cite{dauner2023parting}, improving efficiency compared to traditional RL, which supervises only a single action per rollout. To further alleviate the forgetting issue, R2SE reserves a portion of pretraining objectives that stabilize the process. While performance could be enhanced by incorporating reactive simulation with 3D scene reconstruction and synchronous rollouts, we leave this direction to future work. The overall pipeline is referred to Algorithm~\ref{algo:r2se}.

\subsection{Self-aware Adapter Expansion}
\label{sub3}

While reinforced refinement improves performance on hard cases, it still risks test-time forgetting and overfitting to specialist domains. Prior methods mitigate this via evidential or Bayesian learning~\cite{li2024decode,amini2020deep}, but require front-end pretraining and lack adaptability to hard cases. Instead, R2SE addresses test-time uncertainty from the specialist’s perspective via self-aware adapter expansion. Specifically, it frames test-time forgetting as an out-of-domain (OOD) inference problem~\cite{charpentier2020posterior} based on specialist uncertainty $\mathbf{U}_{\text{test}}= \mathbf{U}(\pi_{\theta+\Delta\theta}(\mathbf{X}))$:
\begin{equation}
\label{equ:ood}
    \mathbf{P}(u) = \int_{u_0}^{u} p_{\hat{\theta}_{\text{Hard}}}(t) \, dt \leq \sigma, \quad u \in \mathbf{U}_{\text{test}},
\end{equation}
where $\mathbf{P}(u)$ denotes the CDF of test-time uncertainty compared to a confidence threshold $\sigma$. The density function $p_{\hat{\theta}_{\text{Hard}}}(t)$ is estimated from hard-case uncertainty values $\mathbf{U}_{\text{Hard}}$ collected from $\mathcal{D}_{\text{Hard}}$, requiring a proper tail-modeling approach. To this end, R2SE leverages the Generalized Pareto Distribution (GPD)~\cite{coles2001introduction}, whose PDF is defined as:
\begin{equation}
\label{equ:gpd}
p_{\text{GPD}}(t) = \frac{1}{\beta} \left(1 + \xi \frac{t - u_0}{\beta} \right)^{-\frac{1}{\xi} - 1}, \quad t \geq u_0,\; \beta > 0,
\end{equation}
with parameters $\hat{\theta}_{\text{Hard}} = (\xi, \beta)$. Supported by~\cite{pickands1975statistical}, GPD is bounded for tail modeling. The parameters can be estimated via maximum likelihood estimation (MLE)~\cite{hosking1987parameter}:
\begin{equation}
\label{equ:mle}
    \hat{\theta}_{\text{Hard}} = \arg\max_{\theta = (\xi, \beta)} \prod_{u \in \mathbf{U}_{\text{Hard}}} p(u \mid \theta, u_0),
\end{equation}
Based on this, R2SE defines the adapter expansion policy:
\begin{equation}\label{equ:expand}
\mathbf{1}(\mathbf{P}_{\text{GPD}}(\mathbf{U}_{\text{test}}) \leq \sigma)\pi_\theta + 
\mathbf{1}(\mathbf{P}_{\text{GPD}}(\mathbf{U}_{\text{test}}) > \sigma)\pi_{\theta+\Delta\theta}.
\end{equation}

That is, if a test case is detected as OOD based on its uncertainty percentile $\mathbf{P}_{\text{GPD}}(\mathbf{U}_{\text{test}})$, R2SE reverts to the generalist policy $\pi_\theta$; otherwise, it expands with the specialist adapter $\pi_{\theta+\Delta\theta}$. Thanks to the ensemble-based adapter design (Sec.~\ref{adapter}), both specialist predictions and uncertainty can be estimated using the mean and variance across $K$ adapters:
\begin{equation}
\begin{aligned}
\label{equ:ens}
    \pi_{\theta+\Delta\theta}(\mathbf{X}) &= \frac{1}{K} \sum_{k=1}^K \pi_{\theta+\Delta\theta_k}(\mathbf{X}), \\
    \mathbf{U}(\pi_{\theta+\Delta\theta}(\mathbf{X})) &= \frac{1}{K} \sum_{k=1}^K \left(\pi_{\theta+\Delta\theta_k}(\mathbf{X}) - \pi_{\theta+\Delta\theta}(\mathbf{X})\right)^2.
\end{aligned}
\end{equation}

Each adapter can also be optionally used for test-time training (TTT), further enhancing robustness. The full adapter expansion process is summarized in Algorithm~\ref{alg:adapter_expansion}.

\begin{algorithm}[t]
\caption{R2SE Adapter Expansion at Test Time}
\label{alg:adapter_expansion}
\begin{algorithmic}[1]
\State \textbf{Input:} Test input $\mathbf{X}$, generalist $\pi_\theta$, specialist $\{\pi_{\theta+\Delta\theta_k}\}_{k=1}^K$, uncertainty threshold $\sigma$
\State \textcolor{gray}{// Precomputed Density using Hard Cases Offline:}
\State Uncertainties $\mathbf{U}_{\text{Hard}}$ from $\mathcal{D}_{\text{Hard}}$ by $\pi_{\theta+\Delta\theta}$ using Equ.~\ref{equ:ens}
\State Estimate GPD $\hat{\theta}_{\text{Hard}} = (\hat{\xi}, \hat{\beta})$ using MLE by Equ.~\ref{equ:gpd},~\ref{equ:mle}
\State \textcolor{gray}{// Test-time Adapter Expansion Steps:}
\State Test-time ensemble $\pi_{\theta+\Delta\theta}(\mathbf{X})$ and $\mathbf{U}(\mathbf{X})$ using Equ.~\ref{equ:ens}

\State Evaluate the GPD CDF score $\mathbf{P}_{\text{GPD}}(\mathbf{U})$ by Equ.~\ref{equ:ood}

\State Switch policy based on Equ.~\ref{equ:expand}:
\If{$\mathbf{P}_{\text{GPD}}(\mathbf{U}(\mathbf{X})) > \sigma$}
    \State Use Specialist policy: $\pi_{\theta+\Delta\theta}$
\Else
    \State Use Generalist policy: $\pi_\theta$
\EndIf
\end{algorithmic}
\end{algorithm}

\section{Experiment}
\label{sec4}
In this section, we first describe the experimental setup for the proposed R2SE framework, including test benchmarks, evaluation metrics, and implementation details. We then present quantitative comparisons against state-of-the-art end-to-end planning methods. To further validate the effectiveness of R2SE, we conduct ablation studies that examine the impact of residual refinement and adapter expansion. Finally, qualitative results are provided to illustrate the advantages of R2SE over existing approaches.
\subsection{Experimental Setup}
\subsubsection{Testing Benchmarks}
To evaluate R2SE in the context of hard-case handling and test-time forgetting, we pose three key questions: (1) How well does R2SE perform in diverse real-world scenarios with model-agnostic E2E autonomous driving systems? (2) Can R2SE improve robustness on standardized hard-case benchmarks and outperform existing methods on challenging scenarios? (3) How effectively does R2SE mitigate forgetting through refinement and expansion? These further guide our benchmarking approach as follows:

\textbf{(1) nuPlan~\cite{dauner2024navsim}:} A large-scale dataset comprising over 700k real-world driving scene clips. We follow the NAVSIM Leaderboard~\cite{navsim_leaderboard}, which R2SE is trained on a curated subset \texttt{navtrain} containing 100k  interactive driving clips, and evaluates on \texttt{navtest}, consisting of 12,146 challenging test scenarios. Each scenario clip is tested through non-reactive simulation with the R2SE deployed for execution.

\textbf{(2) CARLA~\cite{dosovitskiy2017carla}:} A high-fidelity simulation platform with realistic dynamics, suitable for evaluating long-term route planning and scenario-specific driving behaviors. R2SE is assessed using Bench2Drive~\cite{jia2024bench2drive}, a scenario-based benchmark focused on multi-ability behavioral competencies. In this setting,  R2SE is tasked with handling 220 corner-case scenarios defined according to NHTSA Typologies~\cite{najm2007pre}.

\subsubsection{Testing Metrics}
To objectively evaluate the performance of R2SE under E2E configurations, we focus on the final ADS driving performance, assessed using the official NAVSIM leaderboard~\cite{dauner2024navsim} settings. Specifically, in real-world scenarios from nuPlan, R2SE is compared against leaderboard submissions using PDMS as the evaluation metric. Following the same setup as Equ.~\ref{equ:pdms}, the planning trajectories from the E2E model are comprehensively evaluated by PDMS, which encompasses safety (i.e., Non-Collision (\texttt{NC}), Driving Area Compliance (\texttt{DAC}), Time-to-Collision (\texttt{TTC})), Comfort (\texttt{C}), and Expert Progress (\texttt{EP}) against the expert planner operating with privileged ground-truth:
\begin{equation}
\operatorname{PDMS}=\operatorname{NC}\times\operatorname{DAC}\times\frac{5\operatorname{TTC}+5\operatorname{EP}+2\operatorname{C}}{12}.
\end{equation}

For the multi-ability test on Bench2Drive running in CARLA, R2SE is evaluated under the benchmark settings~\cite{jia2024bench2drive} using two metrics: scenario success rate (SR) and the mean Driving Score (DS) across interactive cases. The Driving Score jointly considers Route Completion (RC) and multiplicative infraction scores (IS) as defined in~\cite{dosovitskiy2017carla}:
\begin{equation}
    \operatorname{SR}=\frac{N_{\text{success}}}{N_{\text{total}}};\quad	\operatorname{DS} = \frac{1}{N_{\text{total}}} \sum_{i=1}^{N_{\text{total}}} \operatorname{RC}_i \times \prod_{j=1}^{N_{i,\text{penalty}}} \operatorname{IS}_{i,j}.
\end{equation}

To further evaluate R2SE’s robustness against test-time forgetting, we introduce the standard metric in continual learning called the Forget Rate (FR)~\cite{cao2023continuous}. FR captures the proportion of test cases in which performance declines compared to a previous step:
$
\operatorname{FR} = \frac{1}{N_{\text{total}}} \sum_{i=1}^{N_{\text{total}}} \textbf{1}(\mathbf{M}_i < \mathbf{M}_i^{\text{prev}}),
$ where $\mathbf{M}$ denotes the scalar testing metrics.
Specifically, we define the Hard Improve and Forgetting Rates (HIR, HFR), which represent the proportion of extreme shifts behavior. Moreover, we quantify the Remaining Hard Case to assess R2SE's transferability in tackling test-time hard cases.

\begin{figure}[t]
    \centering
    \vspace{-0.3cm}
    \includegraphics[width=\linewidth]{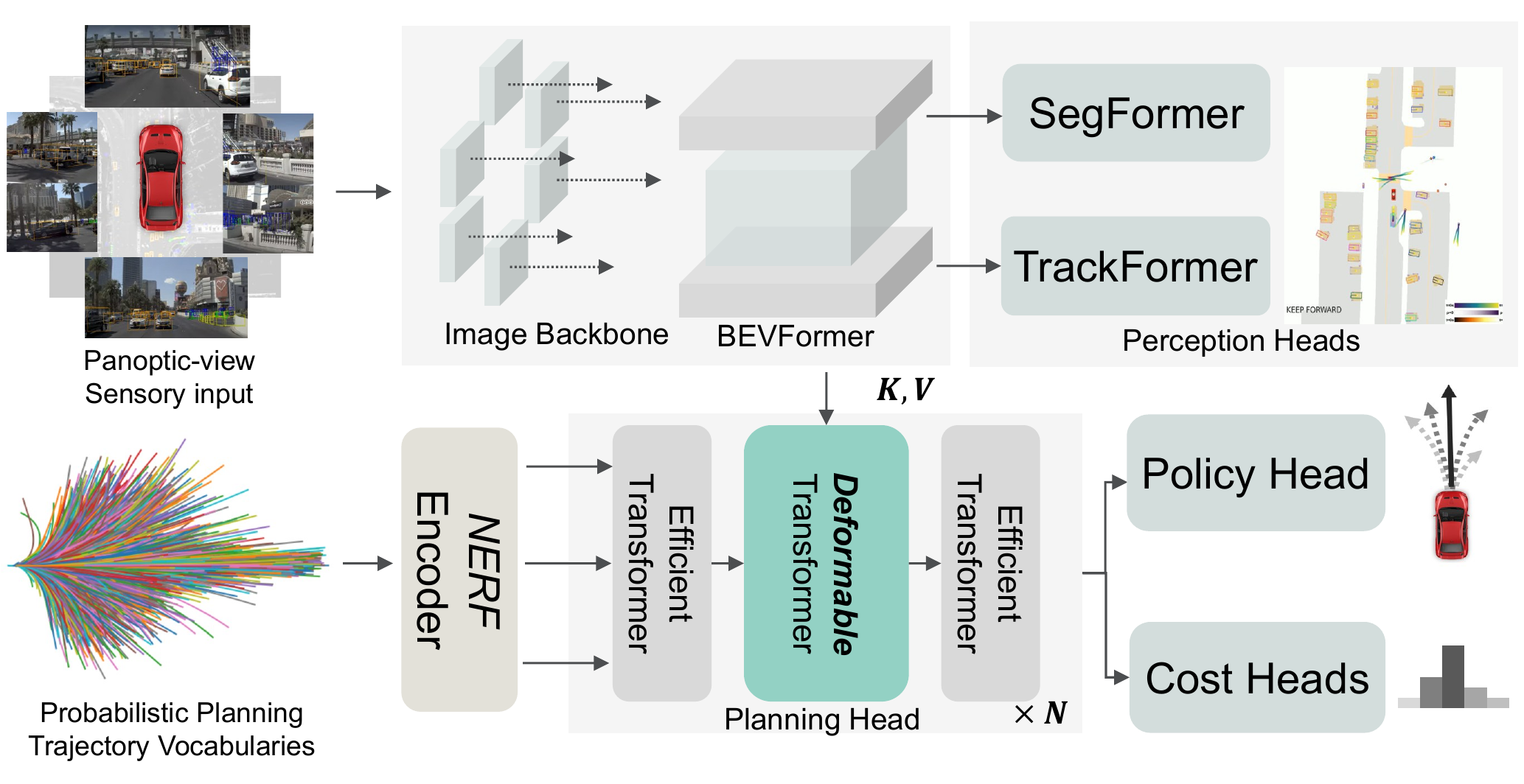}
    \caption[The baseline E2E ADS that serves as the generalist policy.]{The baseline E2E ADS that serves as the generalist policy $\pi_\theta$  in R2SE. A BEV-based sensor backbone extracts features from multi-view sensory inputs $\mathbf{X}$ for actor tracking and mapping~\cite{li2022bevformer}. The planning head operates over planning vocabularies; each planning mode adaptively attends to perception features via deformable attention and outputs $\hat{\mathbf{Y}}$. The planning head can be instantiated using either categorical or diffusion-based variants.}
    \label{fig3}
\end{figure}

\begin{table*}[t]
\caption[Comparison on NAVSIM Leaderboard with closed-loop metrics.]{Comparison on NAVSIM Leaderboard~\cite{navsim_leaderboard} with closed-loop metrics. ($\dagger:$ Pretrained R2SE ADS Base)}
\centering
\small
\setlength{\tabcolsep}{3.8mm}{
 \begin{tabular}{l|c|ccccc|>{\columncolor[gray]{0.9}}c}
    \toprule
    Method & Sensor Input & NC $\uparrow$ &DAC $\uparrow$ & TTC $\uparrow$& Comfort $\uparrow$ & EP $\uparrow$ &  \textbf{PDMS} $\uparrow$  \\
    \midrule
    \textcolor{dgray}{\emph{Human Expert}} & - &  \textcolor{dgray}{\emph{100}} & \textcolor{dgray}{\emph{100}} & \textcolor{dgray}{\emph{100}} & \textcolor{dgray}{\emph{99.9}} & \textcolor{dgray}{\emph{87.5}} & \textcolor{dgray}{\emph{94.8}} \\
    PDM-Closed~\cite{dauner2023parting} & Perception GT &  94.6 & 99.8 & 86.9 & 99.9 & 89.9  & 89.1 \\
    \midrule
    VADv2-$\mathcal{V}_{8192}$~\cite{chen2024vadv2} & Camera \& LiDAR & 97.2 & 89.1 & 91.6 & {100} & 76.0 & 80.9 \\
    Transfuser~\cite{chitta2022transfuser} & Camera \& LiDAR  & 97.7 & 92.8 & 92.8 & {100} & 79.2 & 84.0 \\
    DRAMA~\cite{yuan2024drama} & Camera \& LiDAR  & 98.0 & 93.1 & \underline{94.8} & {100} & 80.1 & 85.5 \\
    Hydra-MDP-$\mathcal{V}_{8192}$-W-EP~\cite{li2024hydra} & Camera \& LiDAR &  \underline{98.3} &96.0 &94.6&{100}& 78.7& 86.5 \\
    GoalFlow~\cite{xing2025goalflow} & Camera \& LiDAR &  98.4  & \underline{98.3}  & 94.6  & 100  & \underline{85.0} & \underline{90.3}\\
    DiffusionDrive~\cite{liao2024diffusiondrive} & Camera \& LiDAR &  98.2  & {96.2}  & 94.7  & {100}  & {82.2}  & 88.1\\
    \midrule
    Hydra-NeXt~\cite{li2025hydra} & Camera  &  {98.1} &{97.7} &{94.6}&{100}& {81.8}& {88.6} \\
    UniAD~\cite{hu2023planning} & Camera & 97.8 & 91.9 & 92.9 & {100} & 78.8 & 83.4 \\
    LTF~\cite{chitta2022transfuser} & Camera & 97.4 & 92.8 & 92.4 & {100} & 79.0 & 83.8 \\
    PARA-Drive~\cite{weng2024drive} & Camera &  97.9 & 92.4 & 93.0 & 99.8 & 79.3 & 84.0 \\
    Hydra-MDP$\dagger$~\cite{li2024hydra}& Camera &  {98.1} & 95.3& 93.9& 100&{81.8}&87.0 \\\midrule
    \textbf{R2SE} & Camera & \textbf{99.0} & \textbf{97.9} & \textbf{96.4} &  \textbf{100} &  \textbf{86.8} & \textbf{91.6} \\
    \bottomrule
\end{tabular}}
\vspace{-0.3cm}
\label{tab:navsim}
\end{table*}
\subsubsection{Implementation Details}
\label{implementation}
To demonstrate the generality of R2SE under agnostic E2E systems, we adopt a minimalist E2E structure, as shown in Fig.~\ref{fig3}. The baseline ADS in R2SE consists of BEV perception~\cite{li2022bevformer} or multi-sensor fusion modules~\cite{chitta2022transfuser}, and a probabilistic planning head. Specifically, R2SE is implemented under both categorical~\cite{li2024hydra,zimmerlin2024hidden}  and diffusion-based~\cite{liao2024diffusiondrive} paradigms, covering two representative SOTA E2E autonomous driving frameworks.

R2SE processes a historical sensory input stream of $T_h = 2$ seconds, with each data frame sampled at 2 Hz. The sensory and perception pipelines follow existing works~\cite{chitta2022transfuser,li2022bevformer}. The planning module operates over a horizon of $T = 4$ seconds in NAVSIM and $T = 2$ seconds in CARLA. The planning query modes $M$ are initialized using clustered logged trajectory vocabularies from the training set: $M = 8192$ for R2SE-MDP and $M = 20$ for the Diffusion setup. A two-layer decoder cascade is used for decoding. Output planning states are passed through the official LQR controller in nuPlan, and similar control setup~\cite{li2025hydra,zimmerlin2024hidden} for executions in CARLA. During GRPO refinement, a distributed~\cite{liang2018rllib} batch simulation platform is designed for $\mathcal{D}_{\text{RL}}$, enabling parallel computation of process rewards and costs.

To ensure fair comparison, R2SE strictly follows the official leaderboard setup, including standardized training and testing. Benchmark results for R2SE are reported using the official metrics. R2SE is trained with a batch size of 16 for categorical setups and 512 for diffusion. It uses the AdamW optimizer with cosine annealing scheduling, and learning rates of $2\mathrm{e}{-4}$ and $6\mathrm{e}{-4}$, respectively. Training and testing are conducted on 8 NVIDIA A100 GPUs. Following Algorithm~\ref{algo:r2se}, hard cases in R2SE are collected offline by scenario logs in nuPlan, and buffered online in CARLA scenarios combining rule-based fallback~\cite{zimmerlin2024hidden} for efficiency,  identified using the 99$^\text{th}$ percentile ($\epsilon =1$) of task difficulty. Expansion criterion of R2SE is controlled by a stricter threshold with $\sigma = 0.75$. Detailed settings are further provided in Appendix~\ref{supp:exp}.

\subsection{Quantitative Results}
\label{sec4:b}
\subsubsection{nuPlan Results}
To validate the real-world driving capability of model-agnostic E2E refinement, two R2SE variants, as detailed in Section~\ref{implementation}, are deployed based on either pure image encoding~\cite{hu2023planning} or sensory fusion with additional LiDAR backbones~\cite{chitta2022transfuser}. Notably, both R2SE variants (R2SE-MDP and R2SE-Diffusion) have achieved leading positions on the {NAVSIM} Leaderboard, surpassing a range of state-of-the-art (SOTA) E2E autonomous driving systems (ADS) characterized by modular integrations~\cite{hu2023planning,jiang2023vad,chitta2022transfuser,yuan2024drama}, probabilistic planning~\cite{chen2024vadv2,li2024hydra}, and planning refinements via diffusion models~\cite{liao2024diffusiondrive,li2025hydra} or goal anchors~\cite{xing2025goalflow}.

Specifically, the R2SE refinement framework boosts expert progress by over $+6.1\%$, enhances safety metrics such as TTC and collision rates by $+2\%$, and leads to a $+3\%$ improvement in PDMS compared to recent SOTA refinement-based ADS~\cite{li2025hydra,liao2024diffusiondrive}. Even larger improvements are observed against prior modular ADS that rely on single-shot planning~\cite{hu2023planning,jiang2023vad,yuan2024drama} or probabilistic distribution modeling~\cite{chen2024vadv2,li2024hydra}, with R2SE reporting a $+13.23\%$ PDMS gain, including a $+5.24\%$ safety improvement and a $+13.02\%$ increase in driving progress~\cite{chen2024vadv2}. Furthermore, R2SE variants also outperform the SOTA rule-based ADS~\cite{dauner2023parting}, which relies on privileged annotations for planning, offering an additional $+10.93\%$ TTC safety improvement and a $+2.5\%$ PDMS increase. Finally, when compared against the founded ADS baseline~\cite{li2024hydra} on Leaderboard, the refinement strategy with expansion in R2SE results in $+5.89\%$ PDMS increasing, which is improved by $2.4\%$ safety enhancement and $+10.29\%$ progress. We refer extended testing results as discussed in Tab.~\ref{tab:navsim_ttt}, Sec~\ref{ttt-content}.

\begin{figure*}[t]
    \centering
    \includegraphics[width=\linewidth]{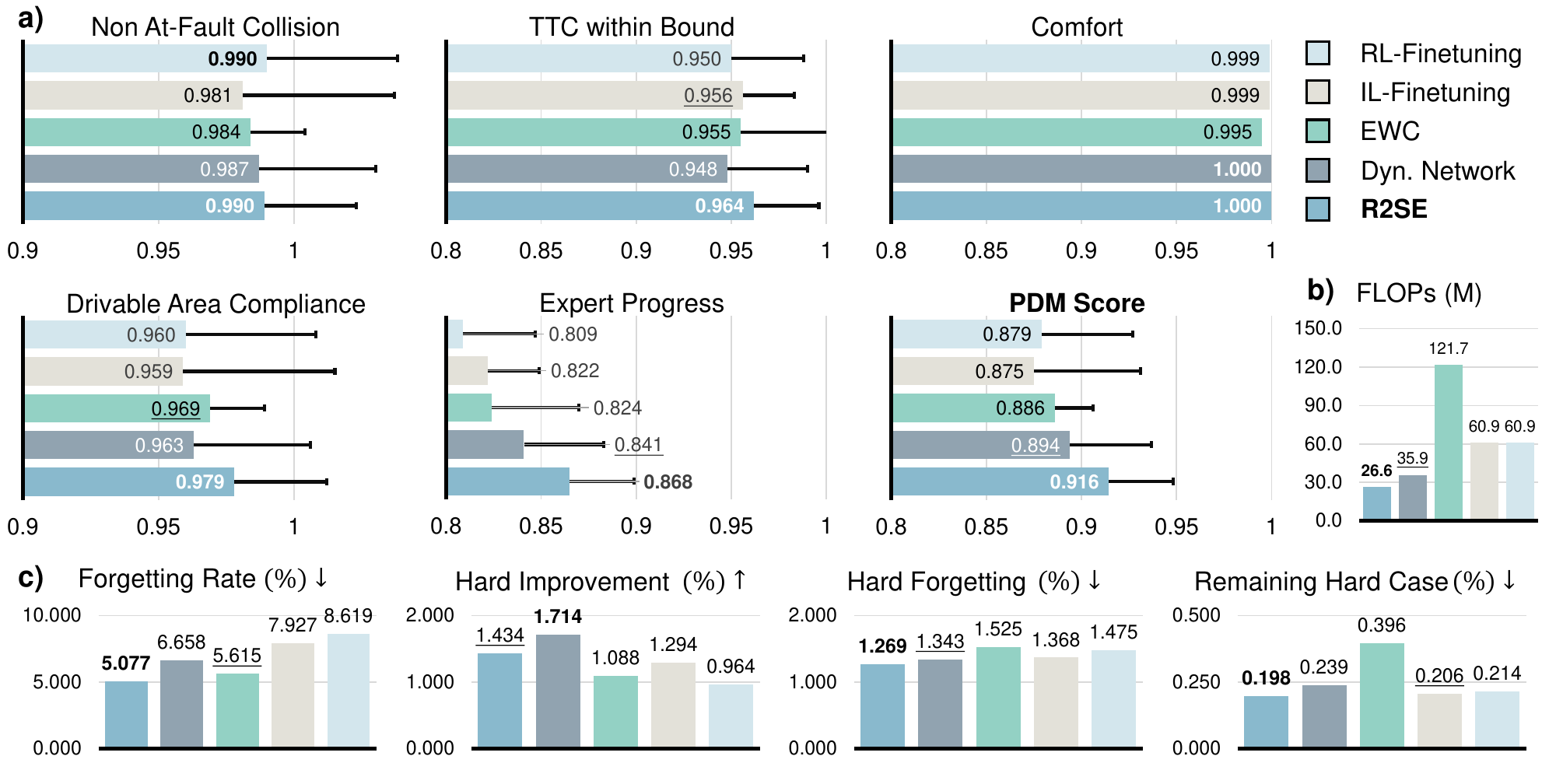}
    \caption[Quantitative comparison across refinement frameworks.]{\textbf{Quantitative comparison across refinement frameworks}. R2SE is evaluated against state-of-the-art refinement paradigms on \texttt{navtest} in terms of (a) general performance metrics, (b) computational cost, and (c) forgetting robustness. (a) R2SE ranks highest in PDMS, achieving strong safety while maintaining driving progress; (b) R2SE and Dynamic Network~\cite{diehl2024lord} exhibit significantly lower computational costs compared to continual learning methods such as EWC~\cite{yang2025human}; (c) full fine-tuning suffers from severe catastrophic forgetting, while EWC and Dynamic Network face challenges in specialized hard case handling and general anti-forgetting, respectively.}
    \label{fig6}
    \vspace{-0.3cm}
\end{figure*}

\begin{table*}[t]
\centering
\caption[Closed-loop and Multi-ability Testing Results of E2E-ADS in CARLA Bench2Drive Leaderboard.]{Closed-loop and Multi-ability Testing Results of E2E-ADS in CARLA Bench2Drive Leaderboard.\\ ($*:$ Privilege Distillation; $\dagger:$ Pretrained R2SE ADS Base)}
\label{tab:bench2drive_full}
\small
\resizebox{\linewidth}{!}{
\begin{tabular}{l|ccc>{\columncolor[gray]{0.9}}c|ccccc>{\columncolor[gray]{0.9}}c}
\toprule
\multirow{2}{*}{\textbf{Method}} & \multicolumn{4}{c|}{\textbf{Closed-loop Metric $\uparrow$}} & \multicolumn{5}{c}{\textbf{Multi-Ability Test} (\%) $\uparrow$} \\
\cmidrule{2-11}
& Efficiency & Comfort& Success& \textbf{DS} & Merging & Overtaking & Emergency Brake & Give Way & Traffic Sign & \textbf{Mean} \\
\midrule
TCP*~\cite{wu2022trajectory} & 54.26 & \underline{47.80} & 15.00 & 40.70 & 16.18 & 20.00 & 20.00 & 10.00 & 6.99 & 14.63 \\
TCP-ctrl*~\cite{wu2022trajectory} & 55.97 & \textbf{51.51} & 7.27 & 30.47 & 10.29 & 4.44 & 10.00 & 10.00 & 6.45 & 8.23 \\
TCP-traj*~\cite{wu2022trajectory} & 76.54 & 18.08 & 30.00 & 59.90 & 8.89 & 24.29 & {51.67} & 40.00 & 46.28 & 34.22 \\
TCP-traj w/o distill.~\cite{wu2022trajectory} & 78.78 & 22.96 & 30.05 & 49.30 & 17.14 & 6.67 & 40.00 & 40.00 & 28.72 & 28.51 \\
ThinkTwice~\cite{jia2023think} & 76.93 & 16.22 & 3.13 & 62.44 & 27.38 & 18.42 & 35.82 & \textbf{50.00} & 54.23 & 37.17 \\
DriveAdapter*~\cite{jia2023driveadapter} & 70.22 & 16.01 & 33.08 & 64.22 & 28.82 & 26.38 & 48.76 & \textbf{50.00} & {56.43} & {42.08} \\\midrule
AD-MLP~\cite{zhai2023rethinking} & 48.45 & 22.63 & 0.00 & 18.05 & 0.00 & 0.00 & 0.00 & 0.00 & 4.35 & 0.87 \\
UniAD-T.~\cite{hu2023planning} & 123.92 & 47.04 & 13.18 & 40.73 & 8.89 & 9.33 & 20.00 & 20.00 & 15.43 & 14.73 \\
UniAD-B.~\cite{hu2023planning} & 129.21 & 43.58 & 16.36 & 45.81 & 14.10 & 17.78 & 21.67 & 10.00 & 14.21 & 15.55 \\
VAD~\cite{jiang2023vad} & 157.94 & 46.01 & 15.00 & 42.35 & 8.11 & 24.44 & 18.64 & 20.00 & 19.15 & 18.07 \\
DriveTransformer-L.~\cite{jia2025drivetransformer} & 100.64 & 46.01 & 35.01 & 63.46 & 17.57 & {35.00} & 48.36 & 40.00 & 52.10 & 38.60 \\
Hydra-NeXt~\cite{li2025hydra}& 197.76& 20.68 & 50.00 & 73.86& 40.00 & 64.44 & 61.67 &\textbf{50.00} & 50.00 & 53.22 \\
TF++~$\dagger$\cite{zimmerlin2024hidden}& \textbf{245.10}& 25.48 & \underline{67.26} & \underline{84.21}& \textbf{58.75} & \underline{57.77} & \underline{83.33} &40.00 & \underline{82.11} & \underline{64.39} \\
\midrule
\textbf{R2SE} & \underline{243.89} & 23.26 & \textbf{69.54} & \textbf{86.28} & \underline{53.33} &\textbf{61.25} & \textbf{90.00} & \textbf{50.00} & \textbf{84.21} & \textbf{67.76} \\
\bottomrule
\end{tabular}}
\vspace{-0.3cm}
\label{tab:carla_b2d}
\end{table*}

\subsubsection{CARLA Results} To manifest the closed-loop driving for E2E systems under standardized hard-case typologies, we evaluate the refinement effect of R2SE against SOTA E2E ADS under the Bench2Drive closed-loop and multi-ability testing protocols. As shown in Table~\ref{tab:carla_b2d}, the R2SE-refined ADS achieves the highest closed-loop driving score and the best average performance in multi-ability testing on the Bench2Drive leaderboard. Specifically, compared to the pretrained baseline ADS~\cite{zimmerlin2024hidden}, R2SE improves the driving score by +2.06 and increases the overall success rate by +2.28. With minimal compromise in efficiency, R2SE consistently improves 4 out of 5 measured driving abilities, achieving a notable safety gain of +6.67 in emergency handling. When compared with other SOTA E2E systems built on diffusion~\cite{li2025hydra} or Transformer-based architectures~\cite{jia2025drivetransformer}, our method delivers solid improvements, yielding a +12.42 driving score increase and a $+27.32\%$ success rate gain in multi-ability tests.

\begin{table*}[t]
\centering
\small
\caption{Roadmap from Hydra-MDP to R2SE-MDP by R2SE algorithmic components}
\setlength{\tabcolsep}{3.7mm}{
\begin{tabular}{l|cccc|ccccc|>{\columncolor[gray]{0.9}}c}
\toprule
ID & IL& GRPO-Cost& GRPO-RL& Expansion& NC $\uparrow$ &DAC $\uparrow$ & TTC $\uparrow$& C. $\uparrow$ & EP $\uparrow$ &  \textbf{PDMS} $\uparrow$  \\\midrule
1 & \textbf{\checkmark}& -& -& - & 98.1 & 95.3& 93.9& 100&81.8&87.0\\
2 & \textbf{\checkmark}& \textbf{\checkmark}& -& -& \underline{98.7} & 96.8& \underline{95.7}& 100&83.8&89.3\\
3 & -& \textbf{\checkmark}& \textbf{\checkmark}& -& \textbf{99.0} & 96.0&\underline{95.7}& 100&80.9&87.9\\
4 & \textbf{\checkmark}& \textbf{\checkmark}& \textbf{\checkmark}& - & \underline{98.7} & \underline{97.0}& 95.5& 100&\underline{84.3}&\underline{89.7}\\
\midrule
0 & \textbf{\checkmark}& \textbf{\checkmark}& \textbf{\checkmark}& \textbf{\checkmark}& \textbf{99.0} & \textbf{97.9} & \textbf{96.4} &  \textbf{100} &  \textbf{86.8} & \textbf{91.6}\\
    \bottomrule
\end{tabular}
}\label{tab:ablation-1}
\vspace{-0.5cm}
\end{table*}

To further evaluate the generality of R2SE across model-agnostic E2E ADS, Tab.~\ref{tab:carla-hard} demonstrates R2SE's synergy with SOTA one-model\cite{wu2022trajectory} and modular~\cite{jiang2023vad} E2E systems. In $20\%$ most difficult Bench2Drive scenarios, R2SE yields solid improvements, achieving +21.31 DS and +29.22 route completions over the original VAD. Similarly, its synergy with TCP results in +31.28 DS and +27.20 completions. These gains indicate R2SE’s effectiveness in improving task completion, particularly in scenarios where the base model stucks, or fails to learn challenging policies. While a marginal increase in infractions may occur (-10.4 in VAD, +8.0 in TCP), it can potentially be mitigated through online guidance.

\subsubsection{Test-time Adaptation results}\label{ttt-content} 
\begin{figure}[t]
    \centering
\includegraphics[width=\linewidth]{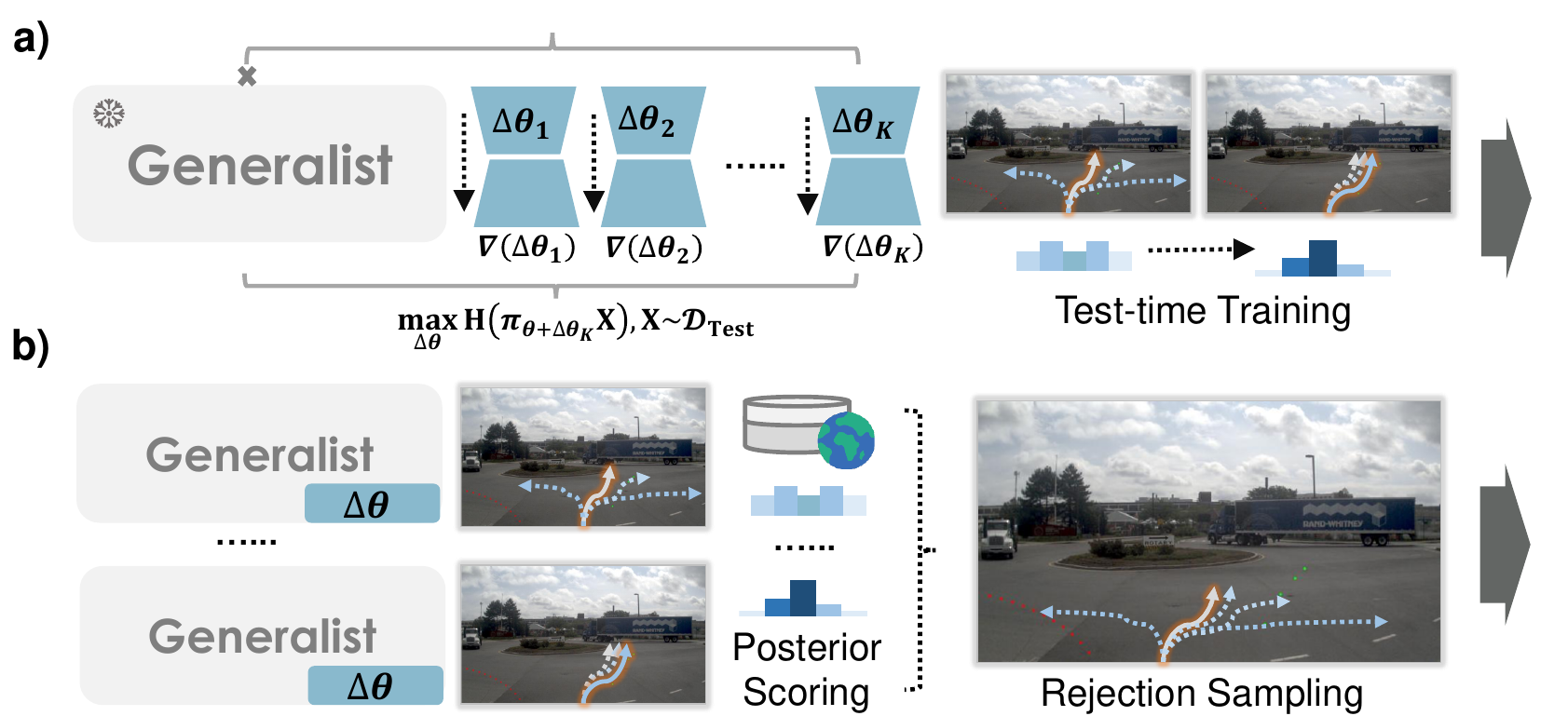}
    \caption[R2SE that seamlessly integrated with online adaptation techniques.]{R2SE could be seamlessly integrated with online adaptation techniques, such as a) Test-Time Training (TTT) or b) Rejection Sampling (RS).}
    \label{fig5}
\end{figure}

The emerging topic of Test-time Adaptation (TTA) and online learning has demonstrated effectiveness in mitigating test-time domain gaps for policy distributions. Leveraging the mixture adapter design for reinforced refinement (Sec.\ref{sub2}), R2SE can be seamlessly integrated with a variety of test-time strategies in an efficient manner. As illustrated in Fig.\ref{fig5}, R2SE could be quickly adapted to \textbf{(a) Test-Time Training (TTT): }~\cite{sima2025centaur}, R2SE fully exploits the adapter parameters to enable batched test-time training. This reduces test-time computational cost while maximizing test-time categorical entropy:
$
    \max_{\Delta \theta} \frac{1}{K}\sum_k^K\mathcal{H}_\text{Cat}\left(\pi_{\theta + \Delta \theta_k}(X)\right),X \sim \mathcal{D}_{\text{Test}}.
$
As in Tab.~\ref{tab:navsim_ttt}, combining TTT~\cite{sima2025centaur} with R2SE without policy expansion yields improved results of +3 PDMS. However, the entropy-based objective leads to suboptimal uncertainty estimation, resulting in comparable PDMS with R2SE-MDP. This indicates uncertainty-aligned loss design for TTT in future work.

\textbf{(b) Rejection Sampling (RS):} Following TrajHF framework~\cite{li2025finetuning}, stacked specialist outputs in R2SE are passed through a posterior PDMS scorer to compute rejection scores. Under identical $\times5$ rejection sampling configuration, as in Tab.~\ref{tab:navsim_ttt}, R2SE achieves a $+2.5$ gain in PDMS and a $+2.6$ improvement in driving progress. This further surpasses the average human performance by $+1.7$ PDMS, thanks to PDMS as posterior for rejection sampling. 

We refer readers to \href{https://huggingface.co/unknownuser6666/R2SE_test/tree/main/R2SE_test}{Per-case Testing Results} for detailed Main Results.

\begin{table}[t]
\centering
\caption{Performance on CARLA Hard Scenarios}
\resizebox{\linewidth}{!}{
\begin{tabular}{lccc|ccc}
\toprule
\textbf{Method} & \multicolumn{3}{c|}{\textbf{Overtaking}} & \multicolumn{3}{c}{\textbf{Aggressive/Accident}} \\
\cmidrule(lr){2-4} \cmidrule(lr){5-7}
& \textbf{DS} $\uparrow$ & \textbf{RC} $\uparrow$ & \textbf{IS} $\uparrow$ & \textbf{DS} $\uparrow$ & \textbf{RC} $\uparrow$ & \textbf{IS} $\uparrow$ \\
\midrule
VAD~\cite{jiang2023vad}         & 28.58 & 54.48 & \textbf{0.668} & \textbf{44.00 }& 58.27 & \textbf{0.769 }\\
\rowcolor{gray!15}\textbf{VAD+R2SE}    & \textbf{47.27} & \textbf{83.70} & 0.564 & 41.22 & \textbf{85.16} & 0.477 \\\midrule
TCP~\cite{wu2022trajectory}         & 15.84 & 49.33 & 0.418 & 23.93 & 74.81 & 0.341 \\
\rowcolor{gray!15}\textbf{TCP+R2SE}    & \textbf{38.24} & \textbf{76.53} & \textbf{0.488} & \textbf{55.21} & 
\textbf{88.72 }& \textbf{0.613} \\
\bottomrule
\end{tabular}
}
\label{tab:carla-hard}
\end{table}

\begin{table}[t]
\centering
\small
\caption[Comparison on Test-time Adaptation Strategies.]{R2SE's Synergy on Test-time Adaptation Strategies ($\dagger:$Test-time Training; $\ddagger:$ PDMS Rejection Sampling $\times5$)}
\resizebox{\linewidth}{!}{
\begin{tabular}{l|ccccc|>{\columncolor[gray]{0.9}}c}
    \toprule
    Method & NC $\uparrow$ &DAC $\uparrow$ & TTC $\uparrow$& C. $\uparrow$ & EP $\uparrow$ &  \textbf{PDMS} $\uparrow$  \\\midrule
    \textcolor{dgray}{MDP-Base}~\cite{li2024hydra} & \textcolor{dgray}{98.1} & \textcolor{dgray}{95.3}& \textcolor{dgray}{93.9}& \textcolor{dgray}{100}&\textcolor{dgray}{81.8}&\textcolor{dgray}{87.0}\\
    Centuar$^\dagger$~\cite{sima2025centaur} & 98.8 & 97.9& 95.9& 100&85.8&90.0\\
    \textbf{R2SE-TTT}$^\dagger$ & \textbf{98.9} &\textbf{ 97.9}& \textbf{96.0}& \textbf{100}&\textbf{85.9}&\textbf{91.3}\\\midrule
    \textcolor{dgray}{Diffusion-Base}~\cite{liao2024diffusiondrive} & \textcolor{dgray}{98.2}  & \textcolor{dgray}{96.2}  & \textcolor{dgray}{94.7}  & \textcolor{dgray}{100}  & \textcolor{dgray}{82.2}  & \textcolor{dgray}{88.1}\\
    TrajHF$^\ddagger$~\cite{li2025finetuning} & 99.3 & \underline{97.5}& \underline{98.0}& 99.8&90.4&94.0\\
TransDiffuser$^\ddagger$~\cite{jiang2025transdiffuser} & \underline{99.4} & 96.5& 97.8& 99.4&\textbf{94.1}&\underline{94.9}\\
    \textbf{R2SE-RS}$^\ddagger$ & \textbf{99.7} & \textbf{99.8}& \textbf{99.3}& \textbf{100}&\underline{92.6}&\textbf{96.5}\\
    \bottomrule
\end{tabular}
}\label{tab:navsim_ttt}
\end{table}

\begin{table}[t]
\centering
\small
\caption{Effect of R2SE tail distribution selection}
\resizebox{\linewidth}{!}{
\begin{tabular}{l|ccccc|>{\columncolor[gray]{0.9}}c}
    \toprule
    $\mathbf{P}(u)$ & NC $\uparrow$ &DAC $\uparrow$ & TTC $\uparrow$& C. $\uparrow$ & EP $\uparrow$ &  \textbf{PDMS} $\uparrow$  \\\midrule
    \textbf{GPD (R2SE)} & \textbf{99.0} & \textbf{97.9} & \textbf{96.4} &  \textbf{100} &  \textbf{86.8} & \textbf{91.6}\\\midrule
    Power Law & {98.8} & \underline{97.7}& \underline{96.0}& 100&\underline{86.2}&\underline{91.1}\\
    Log Normal & \underline{98.9} & 97.4& 95.8& 99.6&84.7&90.2\\
    Percentile & 98.5 & 96.1& 94.9& 100&81.8&87.9\\
    KDE & {98.8} & 97.0& 95.7& 100&83.9&89.6\\
    \bottomrule
\end{tabular}
}
\label{table:hard_dist}
\end{table}

\subsubsection{Refinement Robustness Against Forgetting} To further evaluate the refinement capability of R2SE and its effectiveness in mitigating catastrophic forgetting, we compare R2SE against several state-of-the-art refinement paradigms on the \texttt{navtest} benchmark, as shown in Fig.~\ref{fig6}. The evaluation covers PDMS sub-metrics, including (a) overall performance, (b) efficiency, and (c) forgetting behavior. In terms of general PDMS performance, full fine-tuning strategies~\cite{gao2025rad} based on RL or IL offer only marginal improvements while requiring full computation of the planning head. Although RL-based full tuning achieves the highest non-collision safety score of 0.99, matching R2SE, it suffers from overfitting, resulting in overly conservative behaviors with low driving progress scores of 0.809 and 0.822 in EP. Gradient-based continual learning (EWC)~\cite{yang2025human} improves PDMS to 0.886 by consolidating gradients from prior tasks, constraining overfitting and reducing catastrophic forgetting.

However, its high computational overhead exceeding $\times4.1$ MFLOPs limits its practicality for test-time adaptation. Compared to parameter-isolation approaches like Dynamic Network~\cite{diehl2024lord}, R2SE achieves a $+2.5$ PDMS gain and notable safety improvements. While Dynamic Network mitigates forgetting through parameter isolation and excels in handling hard cases, R2SE's adaptive expansion further enhances robustness, particularly when adapter modules exhibit uncertainty.

Fig.~\ref{fig6}c further quantifies forgetting robustness. R2SE achieves the lowest forgetting rate and the fewest remaining hard cases on \texttt{navtest}, reducing case forgetting by $-41\%$ and improving hard case handling by $+48.7\%$ relative to RL full tuning, reinforcing our earlier observations of severe forgetting in full tuning. EWC performs well in general forgetting, achieving a $+19.5\%$ improvement in hard case handling over R2SE; however, its difficulty in adapting to sparse corner cases sacrifices specialist performance, evidenced by its highest remaining hard case rate (0.396) and elevated forgetting rate on hard cases. Contrarily, Dynamic Network specializes in corner case, achieving the highest hard case improvement (1.714) but less effectiveness in general forgetting ($-31.1\%$), whereas R2SE maintains better overall robustness through uncertainty-aware expansion.

\begin{figure*}[t]
    \centering
    \includegraphics[width=\linewidth]{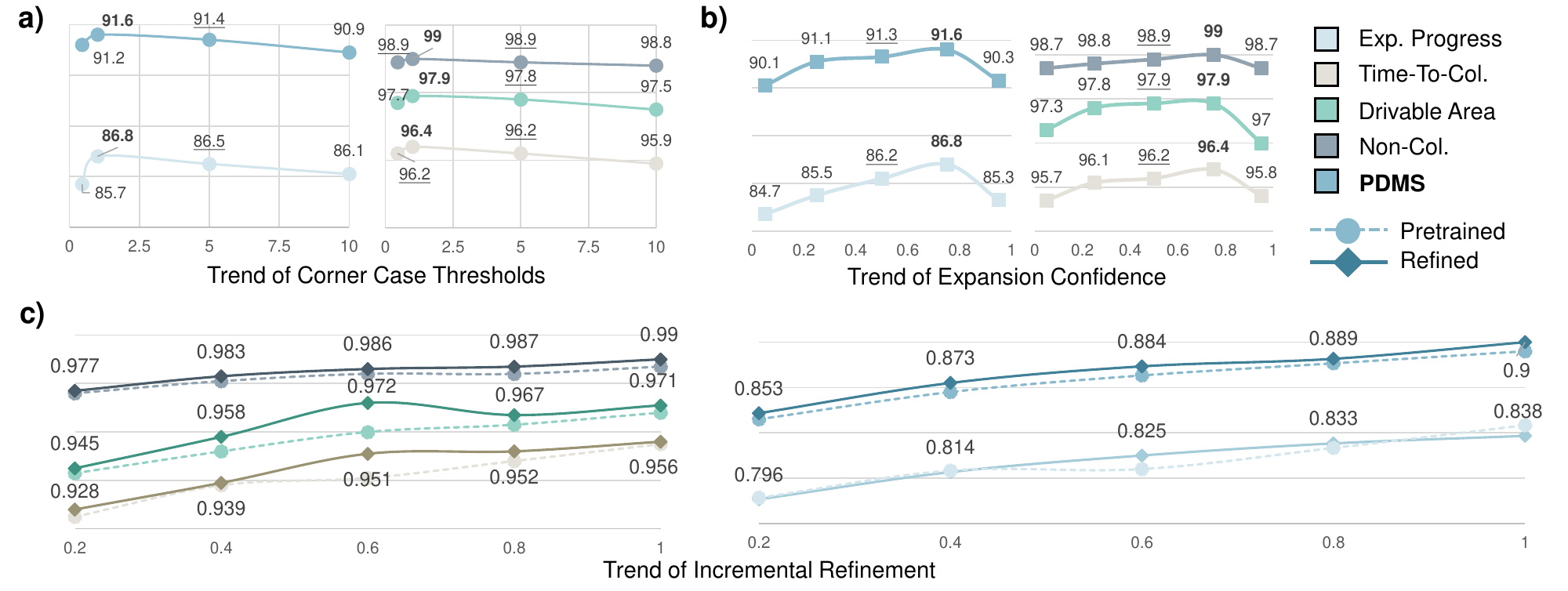}
    \caption[Ablation studies for R2SE trends.]{\textbf{Ablation studies for R2SE trends}. a) Convergence effect of increasing corner case threshold $\epsilon$ for loosing difficulty criteria. b) A reversed converging trend with stricter expansion confidence $\sigma$. c) R2SE consistently improves E2E driving under incremental refinement by \texttt{navtrain} subset splits.}
    \label{fig8}
    \vspace{-0.3cm}
\end{figure*}

\subsection{Ablation Studies}
\label{sec4:c}
To delve into the core mechanisms of each component within the R2SE framework, targeted ablation studies are conducted on \texttt{navtest}. Leveraging the base of \textbf{R2SE-MDP}, specific effects are discussed  corresponding to the reinforced refinement and self-aware expansion modules.

 \begin{table}[t]
\centering
\small
\caption{Effect of R2SE hard case threshold}
\resizebox{\linewidth}{!}{
\begin{tabular}{l|ccccc|c}
    \toprule
     Case Difficulty $\epsilon$ & NC $\uparrow$ &DAC $\uparrow$ & TTC $\uparrow$& C. $\uparrow$ & EP $\uparrow$ &  \textbf{PDMS} $\uparrow$  \\\midrule
    \rowcolor{gray!0}0.45 ($\texttt{PDMS}=0$)& \underline{98.9}& 97.7& \underline{96.2}& 100&85.7&91.2\\
    \rowcolor{gray!10}\textbf{1.00 (R2SE)}& \textbf{99.0} & \textbf{97.9} & \textbf{96.4} &  \textbf{100} &  \textbf{86.8} & \textbf{91.6}\\
    \rowcolor{gray!20}{5.00}& \underline{98.9} & \underline{97.8} & \underline{96.2} &  {100} &  \underline{86.5} & \underline{91.4}\\
    \rowcolor{gray!30}10.0 & 98.8& 97.5& 95.9& 100&86.1&90.9\\
    \bottomrule
\end{tabular}
}
\label{table:hard_case_thres}
\end{table}

\begin{table}[t]
\centering
\small
\caption{Effect of R2SE expansion confidence}
\resizebox{\linewidth}{!}{
\begin{tabular}{l|c|ccccc|c}
    \toprule
    Confidence $\sigma$&ER & NC $\uparrow$ &DAC $\uparrow$ & TTC $\uparrow$& C. $\uparrow$ & EP $\uparrow$ &  \textbf{PDMS} $\uparrow$  \\\midrule
    0.05 & 2.09&98.7 & 97.3& 95.7& 100&84.7&90.1\\
    \rowcolor{gray!10}0.25& 10.1& \underline{98.8} & \underline{97.8}& {96.1}& 100&85.5&91.1\\
    \rowcolor{gray!20}0.50& 25.9& \textbf{98.9} & \textbf{97.9}& \underline{96.2}& 100&\underline{86.2}&\underline{91.3}\\
    \rowcolor{gray!30}\textbf{0.75 (R2SE)}& 50.0&\textbf{99.0} & \textbf{97.9} & \textbf{96.4} &  \textbf{100} &  \textbf{86.8} & \textbf{91.6}\\
    \rowcolor{gray!35}0.95&82.8& 98.7 & 97.0& 95.8& 100&85.3&90.3\\
    \bottomrule
\end{tabular}
}
\label{tab:confidence_thres}
\end{table}

\subsubsection{Effect of R2SE Refinement Components} To examine the contribution of each algorithmic component in R2SE, Table~\ref{tab:ablation-1} presents an ablative roadmap from the original HydraMDP baseline~\cite{li2024hydra} (ID-1) to the fully refined R2SE-MDP (ID-0). First, introducing cost constraints (Equ.~\ref{equ:constraint}) plays a critical role in ensuring effective refinements (ID-1 vs. ID-2). Soft constraints guide the offline exploration in GRPO toward safer and more efficient trajectories, yielding a gain of $+2.3$ in PDMS, with improvements of $+2.0$ in driving progress and $+1.8$ in TTC. This coincides with recent discussion in~\cite{gao2025rad,zhang2025carplanner}, as RL spends costly data consumptions against policy degradation~\cite{huang2024gen} with random exploration. In contrast, performing reinforcement learning fine-tuning without an imitation learning (IL) objective (ID-1 vs. ID-3) yields marginal improvement (+0.9 PDMS), despite targeted optimizations for safety and cost. While this full objective leads to state-of-the-art safety scores (99.0 in NOC and 95.7 in TTC), the absence of expert guidance causes the specialist modules to overfit, overriding the generalist policies learned via IL, and leading to catastrophic forgetting. Reintroducing the IL objective (ID-3 vs. ID-4) improves PDMS by +1.8, driven largely by a +4.3 increase in progress, with minimal sacrifice in safety. This indicates that while IL aids generalization, it cannot alone determine whether to rely on the generalist or specialist in varying scenarios. Finally, incorporating the expansion mechanism in R2SE (ID-4 vs. ID-0) yields a further gain of +1.9 in PDMS and +2.5 in progress, demonstrating advanced performance across all metrics through self-aware adaptation to both general and hard cases.

\begin{figure}[t]
    \centering
    \includegraphics[width=\linewidth]{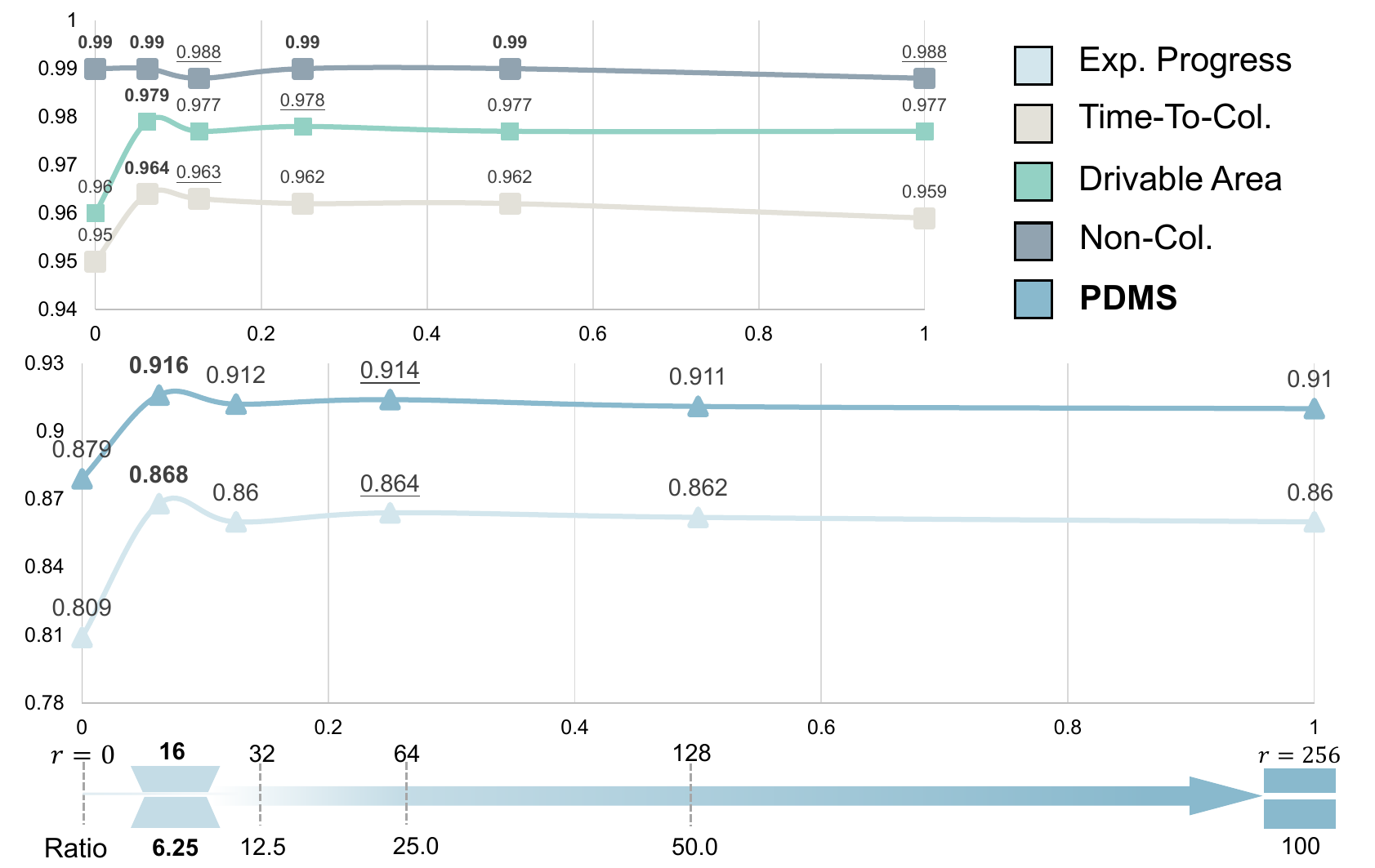}
    \caption[Effect of Specialist Adapter Rank.]{\textbf{Effect of Specialist Adapter Rank}.R2SE reports fluctuated convergence by increasing adapter ranks.}
    \label{fig12}
\end{figure}

\begin{figure*}[t]
    \centering
    \includegraphics[width=\linewidth]{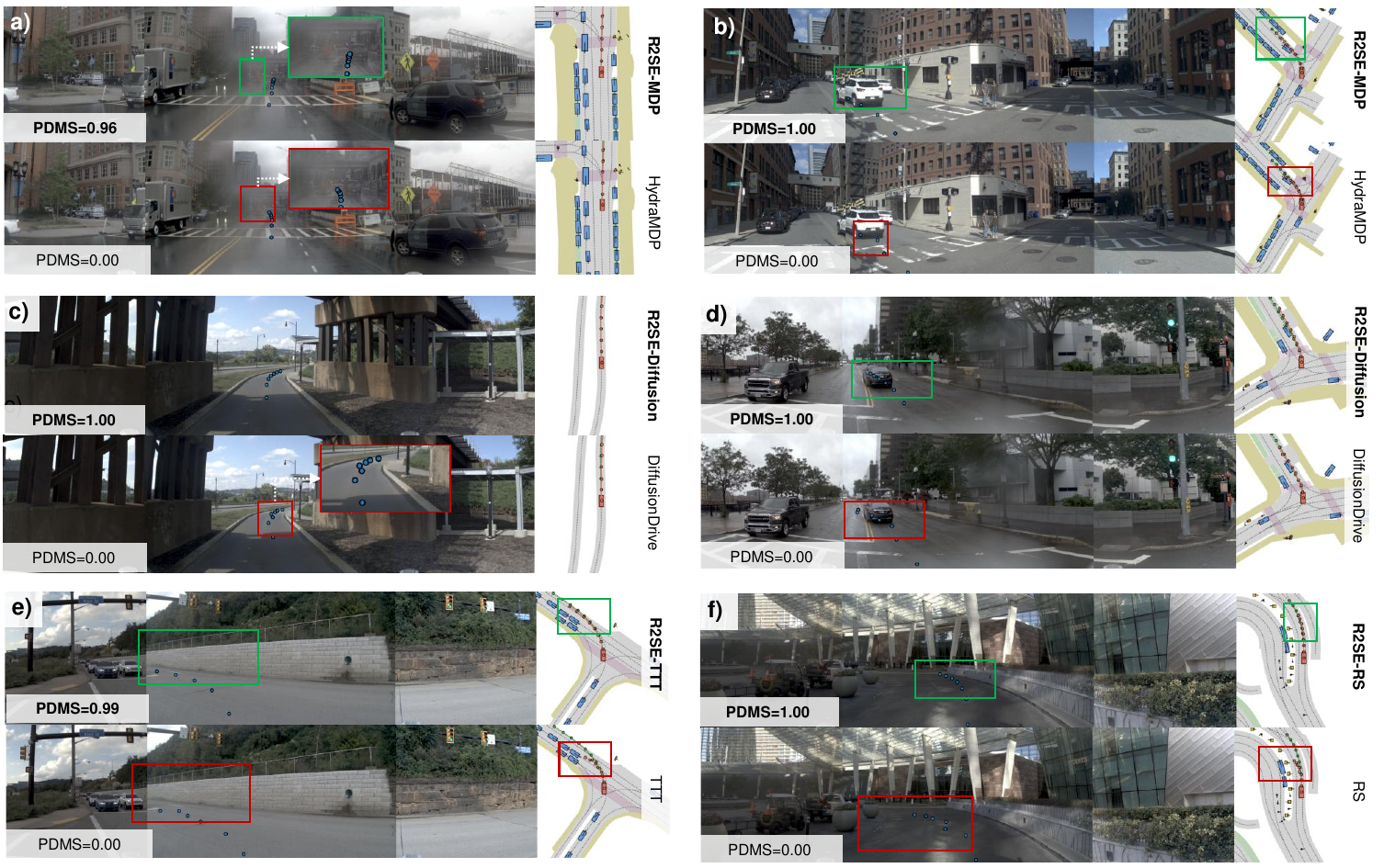}
    \caption[Qualitative results on navsafe test splits.]{\textbf{Qualitative results on} \texttt{navsafe} \textbf{test splits}. Model-agnostic R2SE is evaluated against SOTA E2E ADS on nuPlan under NHTSA Typologies.
a)-b): Compared to HydraMDP~\cite{li2024hydra}, the categorical form of R2SE-MDP can a) more accurately identify under blurred sensory conditions, and b) moderately correction during unprotected turns.
c)-d): Compared to the diffusion-based ADS~\cite{liao2024diffusiondrive}, R2SE-Diffusion provides better denoising refinements for c) rural roads with shaded bridges, and d) off-route turns across double yellow lines caused by occlusions.
(e)-(f) R2SE also demonstrates synergy with online adaptations: (e) recovering from TTT~\cite{sima2025centaur} failures, and (f)  smoothing planning when regular rejection sampling~\cite{li2025finetuning} misidentifies left-hand-side blobs.}
    \label{fig7}
    \vspace{-0.3cm}
\end{figure*}

\subsubsection{Effect of Adapter Ranks} We further examine R2SE specialists by varying the ratio of adapter rank $r$, to assess the impact of adapter-induced specialist parameter ratios. As shown in Fig.~\ref{fig12}, transitioning from non-adapter finetuning to full-rank adapter tuning reveals a non-monotonic convergence pattern. This fluctuation arises from the instability associated with parameter plasticity and forgetting. As further discussed in~\cite{biderman2024lora}, increasing the adapter rank tends to amplify the likelihood of overwriting previously learned general knowledge, leading to potential degradation in generalization.

\subsubsection{Effect of Hard Case Threshold} To further investigate the impact of hard case quantity on the refinement capability of R2SE, Table~\ref{table:hard_case_thres} reports performance variations when hard cases are collected using different difficulty thresholds $\epsilon$. The refined ADS demonstrates a converging trend (Fig.~\ref{fig8} a) in PDMS, showcasing strong refinements (91.2 PDMS, a gain of +4.7) using as few as 500 hard training cases with PDMS = 0 ($0.45\%$ of the training data). R2SE achieves peak performance when leveraging hard cases defined by the top $\epsilon=1$ percentile. Increasing $\epsilon$ introduces more hard samples and potentially enables better estimation of the distribution boundary for expansion, supported by the Central Limit Theorem (CLT) as discussed in~\cite{cao2021confidence,cao2023continuous}. However, loosening the difficulty threshold may introduce incremental sub-hard cases that lack clear distributional boundaries, thereby reducing the accuracy of distributional inference. As a result, an overly large threshold can compromise performance, as reflected by a PDMS drop of –0.7 under a looser selection criterion ($\epsilon = 10$).


\subsubsection{Effect of Expansion Distribution}To evaluate R2SE’s expansion selection in modeling tail distributions, we compared its performance using a range of common statistical distributions~\cite{coles2001introduction}, with particular emphasis on the Generalized Pareto Distribution (GPD). As shown in Table~\ref{table:hard_dist}, GPD yields the most effective inference for determining whether test cases fall within the specialist domain. This aligns with Extreme Value Theory (EVT). As proved in Theorem~\ref{theorem2}, the distribution of excesses over a high threshold converges to a GPD under mild regularity conditions, making it asymptotically optimal for tail approximation. As a special case of GPD, the power-law distribution ranks second, incurring a moderate performance drop of –0.5 PDMS and –0.6 EP. This is expected since tail behaviors can be either exponential ($\xi<0$) or bounded ($\xi=0$), and GPD provides greater flexibility to capture such variations. In contrast, distributions from the normal family, such as the log-normal or kernel density estimation (KDE), tend to oversmooth or underestimate tail risks. KDE, in particular, is highly sensitive to bandwidth selection and performs poorly (–2.0 PDMS), even degrading performance relative to the original RL-based refinement. Similarly, percentile-based methods exhibit substantial declines (–5.0 EP and –5.7 PDMS), further underscoring the necessity of proper tail modeling in expansion selection.

\subsubsection{Effect of Expansion Confidence} To investigate the impact of confidence threshold $\sigma$ in gating specialist policy activation via GPD, we evaluate R2SE's test-time performance under increasingly strict confidence levels. As shown in Table~\ref{tab:confidence_thres}, a higher $\sigma$ leads to more frequent fallback to the generalist policy, as reflected by the rising Expand Rate (ER). This fallback behavior causes a non-monotonic performance trend: while modest increases in $\sigma$ (e.g., $\sigma = 0.05$) lead to a mild PDMS drop of $-1.5$, overly strict thresholds (e.g., $\sigma = 0.95$) result in a substantial degradation of $-1.3$ PDMS, as shown in Fig.~\ref{fig8}b. This aligns with the intuition that hard cases become increasingly difficult to confidently bound under loose criteria, and thus are either rejected or handed off to the generalist. Therefore, a moderately strict threshold (e.g., $\sigma = 0.75$) achieves the best overall driving performance. In contrast, overly conservative thresholds may discard correctly identified hard cases, degrading the specialist’s contribution and reducing the system to generalist behavior.

\begin{figure*}[t]
    \centering
    \includegraphics[width=\linewidth]{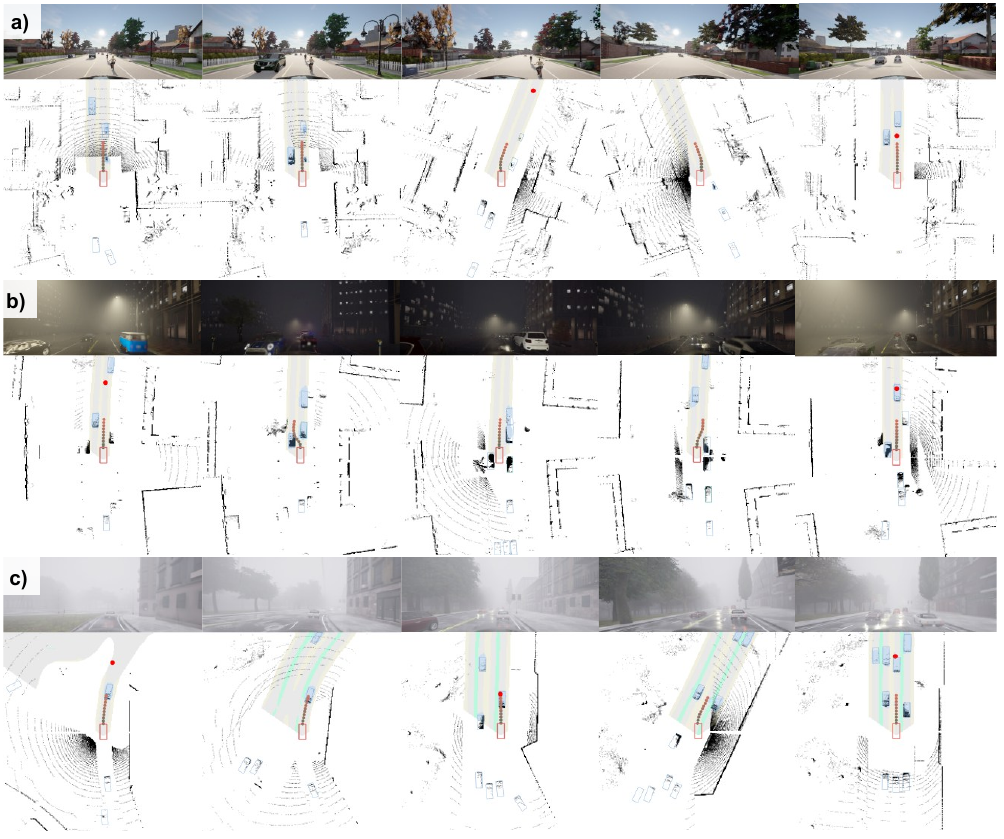}
    \caption[Qualitative Results on Bench2Drive test scenarios.]{\textbf{Qualitative Results on Bench2Drive test scenarios.} a) R2SE-refined ADS identifies appropriate nudging windows to overtake cyclist flows. b) At night, the ADS performs an overtaking maneuver by temporarily entering the opposite lane upon detecting construction barriers. c) In hazy weather, the ADS successfully merges from the ramp into the fast lane.}
    \label{fig11}
    \vspace{-0.3cm}
\end{figure*}

\subsubsection{Effect of Incremental Refinements} Another important aspect is evaluating R2SE’s refinement capability under incremental (continuous) data streams. As illustrated in Fig.~\ref{fig8}c, we randomly partition the \texttt{navtrain} dataset into five sub-splits to simulate incremental learning setting. Compared to the static pretrained model, R2SE consistently achieves progressive performance improvements across each stage. Notably, it also accelerates convergence toward optimal driving performance, boosting around 50\% in data efficiencies. This further presents R2SE's potential in  continuously evolving refinements.

\subsection{Qualitative Results}
To highlight the refinement brought by R2SE under challenging scenarios, we qualitatively assess E2E performance firstly on \texttt{navsafe}\cite{sima2025centaur}, a hard-case subset of \texttt{navtest} annotated by human experts according to the NHTSA Pre-Crash Typologies\cite{najm2007pre}. As shown in Fig.~\ref{fig7}, R2SE consistently improves base E2E ADS across a variety of difficult settings in a model-agnostic manner. Refinements guided by GRPO more effectively distinguish correct policy behaviors when perception inputs are degraded by extreme weather, sensory smudging (Fig.\ref{fig7} a), or challenged by highly interactive scenarios, such as dense traffic or unprotected turns (Fig.\ref{fig7} b). Additionally, R2SE facilitates structural denoising: in shaded environments (Fig.\ref{fig7} c) or occluded topologies (Fig.\ref{fig7} d), it enables the model to recover consistent road geometry, resulting in smoother and safer planning. Moreover, R2SE demonstrates robustness where prior online methods falter, successfully recovering actionable trajectories under sensory corruption or complex agent interactions (Fig.\ref{fig7} e, f). 

\begin{figure*}[t]
    \centering
    \includegraphics[width=\linewidth]{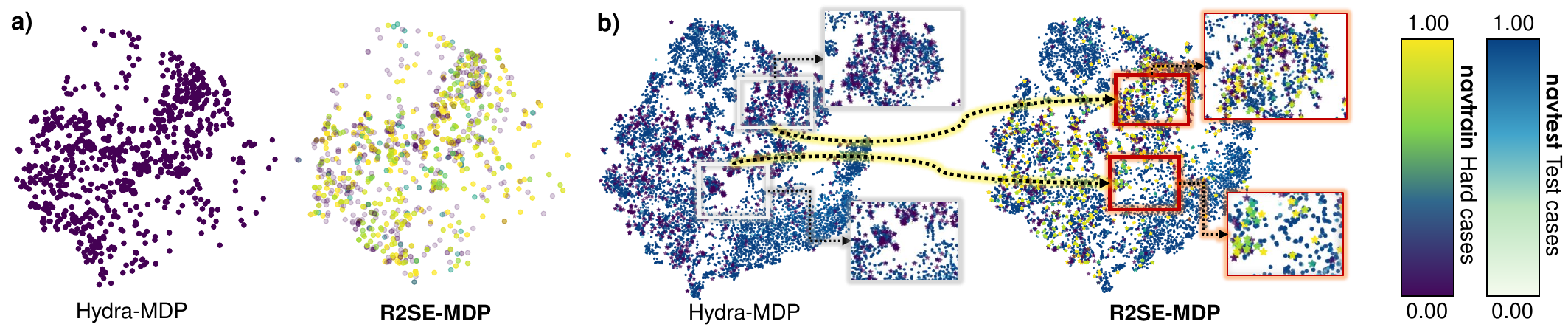}
    \caption[Qualitative Results on Hard Case Improvement by R2SE.]{\textbf{Qualitative Results on Hard Case Improvement by R2SE.} t-SNE visualizations colored by corresponding PDMS are presented for \texttt{navtrain} hard cases and \texttt{navtest} test cases. a) R2SE largely improves PDMS on training hard cases with minimal displacement in feature space. b) The refined training hard cases exhibit improved transferability to the test domain, resulting in a more uniformly distributed and balanced t-SNE embedding for test cases.}
    \label{fig9}
    \vspace{-0.3cm}
\end{figure*}

\begin{figure}[t]
    \centering
    \includegraphics[width=\linewidth]{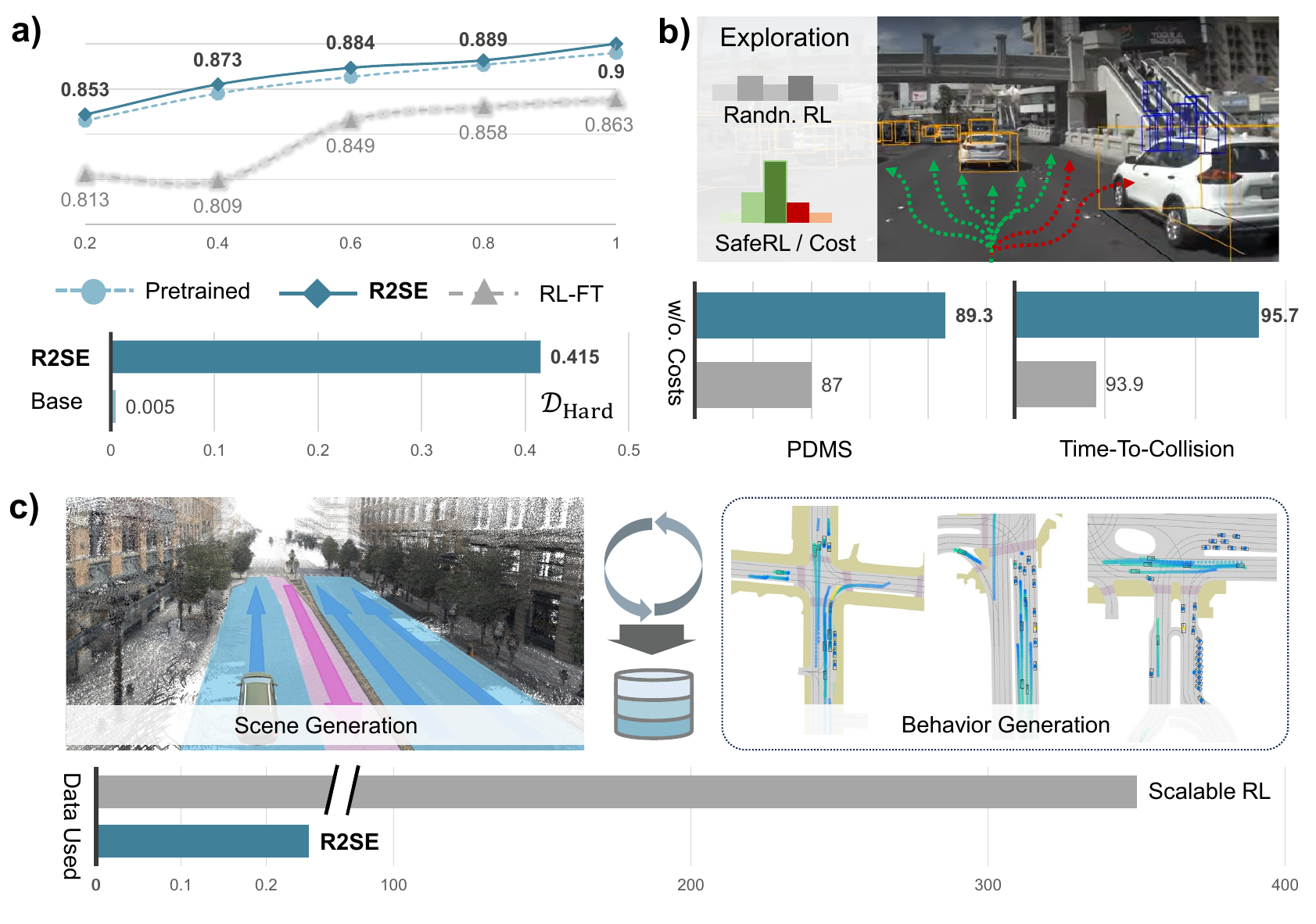}
    \caption[Challenge and frontier upon R2SE.]{\textbf{Challenge and frontier upon R2SE.} a) Combat with catastrophic forgetting of RL finetuning (RL-FT); b) Cost / Safe RL explorations are necessary in E2E ADS; c) Data efficiency matters in R2SE compared with scalable RL~\cite{cusumano2025robust}, and could be further boosted by E2E generative learning. (Scene figures refer to~\cite{li2025mtgs,zhou2025decoupled}.)}
    \label{fig10}
\end{figure}

We further demonstrate the effectiveness of R2SE through qualitative results on Bench2Drive, showcasing its performance on previously underperforming hard cases encountered by the baseline. As illustrated in Fig.\ref{fig11}, R2SE robustly handles diverse challenging scenarios, including cyclist nudging (Fig.\ref{fig11} a), opposite-lane overtaking (Fig.\ref{fig11} b), and ramp merging (Fig.\ref{fig11} c), across a range of weather conditions.

To further underscore R2SE’s capacity for enhancing safety-critical driving and its transferability to broader domains, we perform t-SNE visualization on the embedded space of hard case samples $\mathcal{D}_{\text{Hard}}$ collected from \texttt{navtrain} (used for fine-tuning) and test cases from \texttt{navtest}, each annotated with their respective PDMS scores. Fig.\ref{fig9} illustrates the resulting t-SNE distributions. Qualitatively, the targeted refinement yields clear improvements in the training hard cases, as shown in Fig.\ref{fig9} a, while preserving a stable feature representation with minimal distortion, thanks to the specialist adapter designed in R2SE. Furthermore, when visualizing the joint embedding of $\mathcal{D}_{\text{Hard}}$ and test cases, we observe that compared to original Hydra-MDP, R2SE-refined hard cases exhibit better generalization to the test domain (Fig.~\ref{fig9} b), manifesting as a more uniformly distributed cluster of high-scoring test cases. This demonstrates R2SE’s effective transfer of specialized refinement through its expansion mechanism, which promotes broader generalizability.

\subsection{Challenge and Frontier}
The Era of Experience~\cite{silver2025welcome} unfolds by scaled RL deployments. While R2SE achieves SOTA results (Sec.\ref{sec4:b}) and its mechanisms are discussed in Sec.\ref{sec4:c}, this section highlights emerging patterns observed across reinforced frameworks, offering future directions upon R2SE.

\subsubsection{Anti-Forgetting}
R2SE effectively addresses hard cases during training, with substantial gains in NAVSIM (Fig.\ref{fig9} a) and CARLA (Tab.\ref{tab:carla-hard}) that generalize to test-time scenarios (Fig.~\ref{fig9} b). It serves as a surrogate for domain robustness via residual refinement and expansion, mitigating forgetting (Fig.\ref{fig6}), aided by low-rank adapters that learn and forget less\cite{biderman2024lora}. Fig.~\ref{fig10} a demonstrates catastrophic forgetting by continual RL finetuning (RLFT), reporting -3.1 to -7.3 PDMS, while R2SE consistently improves test scenarios and training hard cases (+41 PDMS). Future work will extend R2SE with domain generalization and continual adaptation. Integrating evidential learning may further enable adapters to internalize statistical expansion. Ultimately, R2SE aims to mimic human-like learning, continually evolving through active learning and selective forgetting.

\subsubsection{Constraints are Necessary}
Safe exploration, via cost constraints or prediction, is critical for RL in E2E ADS. While scalable RL has shown impressive results ~\cite{cusumano2025robust}, it is vulnerable to uncontrolled behaviors misaligned with real-world expectations, especially when diverging from human-like patterns. For deployable E2E ADS, demonstration-guided learning remains essential, supplemented by structured, cost-aware exploration to avoid uninformative behaviors, as exemplified by R2SE in Tab.~\ref{tab:ablation-1} and Fig.~\ref{fig10} b.  RL alone is not a silver bullet: practical deployment hinges more on effective cost prediction, as exemplified by~\cite{gao2025rad}, providing densely aligned and informative reinforced signals.

\subsubsection{Data Efficiency Matters} While RL in LLMs mainly boosts data efficiency rather than reasoning~\cite{yue2025does}, its long-term value in E2E autonomous driving lies in leveraging feedback-driven exploration rather than conventional value learning. As demonstrated in R2SE, the key challenge is to use RL as a semi-supervised signal to augment OOD cases beyond demonstrations. Currently, R2SE tunes only on known hard cases. As denoted in Fig.~\ref{fig10} c, given around $1400\times$ data requirements for scalable RL, future work will incorporate generative frameworks for scene reconstruction~\cite{li2025mtgs} and agent behavior~\cite{zhou2025decoupled}, to expand hard-case coverage, enhancing data efficiency while promoting scalable self-learning with greater interpretability and controllability.

\section{Conclusion}
\label{sec5}
We presented R2SE, a reinforced refinement framework for model-agnostic end-to-end (E2E) autonomous driving systems. Built upon arbitrary pretrained generalist policies, R2SE identifies and addresses residual hard cases through targeted case allocation and reinforced residual learning for specialist modules. By selectively expanding confident specialist behaviors during inference, R2SE enables effective test-time adaptation without compromising the integrity of the generalist policy. Extensive testing on nuPlan, CARLA, and safety-critical subsets aligned with NHTSA Typologies demonstrate that R2SE achieves state-of-the-art refinements across a wide range of E2E systems. Moreover, R2SE exhibits strong synergy with various online adaptation frameworks and SOTA robustness against catastrophic forgetting compared with diverse refinement methods. Ablative results further reveal the internal mechanisms of R2SE. These findings highlight R2SE’s potential as a scalable solution for hard case generalization and deployment in E2E ADS. Future work may extend R2SE with generative frameworks to further improve E2E data efficiency and strengthen anti-forgetting capabilities.

\bibliographystyle{IEEEtran}
\bibliography{b1}

\appendix
\subsection{Additional Testing Results}\label{supp:test_results} We supplement detailed \href{https://huggingface.co/unknownuser6666/R2SE_test/tree/main/R2SE_test}{Per-case Testing Results} of R2SE refined E2E systems on NAVSIM~\cite{navsim_leaderboard} (nuPlan) and Bench2Drive~\cite{jia2024bench2drive} (CARLA) benchmarks.
\subsection{Notation and Setup}\label{supp:exp} 
We provide additional notations and their corresponding values for the experimental platform as follows:
\begin{table}[htp]
\caption{{Notations and parameters}}
\centering
\resizebox{\linewidth}{!}{
\begin{tabular}{llll}
\toprule
\multirow{2}{*}{Notation}& \multirow{2}{*}{Meaning}            & \multicolumn{2}{c}{Testing Benchmarks}  \\
                        &                                      & nuPlan & CARLA       \\ \midrule
$H,W$                   & Size of BEV space                    & 200     & 128         \\
                        & BEV length (m)                       & 50      & 64          \\
        &Image Resolutions& 960$\times$540 & 1024$\times$512\\
        &LiDAR Resolutions& - & 256$\times$256\\
$T$                     & Future Horizons (s)                  & 4       & 2           \\
$T_h$                   & Historical Horizons (s)              & 2     & 0           \\
                        & Temporal frequency (Hz)              & 2       & 2           \\
$M$                     & Number of modalities                 & 8192 & 8    
\\
$N$                     & Planner decoder layers                  & 3       & 3            \\\midrule
$\beta_{\text{Per}}$    & Perception hard case weight          & 0.1     & 0.1          \\
$\beta_{\text{Ent}}$   & Entropy hard case weight             & 0.01    & 0.01         \\
$\epsilon$              & Hard case threshold                  & 1.00    & 1.00        \\
$K$                     & Adapter ensemble numbers             & 6       & 6            \\
$r$                     & Adapter ranks                        & 16      & 16           \\
$\gamma$     &  Decay weight        & 0.99     & 0.99          \\
$\alpha$     & Planning  weight     & 1     & 1          \\
$\lambda$     & Cost  weight     & 1     & 1          \\
$d$                     & Embedding dimensions                 & 256     & 256          \\
$\sigma$                & Expansion confidence                 & 0.75    & 0.75         \\\midrule
                        & Activation function                  & \multicolumn{2}{c}{$\operatorname{ReLU}$}\\
                        & Dropout rate                         & 0.1     & 0.1          \\
                        & Batch size                           & 32       & 32     \\
                        & Learning rate                        & $1e^{-4}$ & $1e^{-4}$    \\
                        & Training epochs                      & 8      & 30 \\ \bottomrule
\end{tabular}
}\vspace{-0.3cm}
\label{table:notation}
\end{table}

\begin{table}[t]
\caption{{Additional Notations}}
\centering
\setlength{\tabcolsep}{5mm}{
\begin{tabular}{ll}
\toprule
{Notation}& {Meaning}   \\\midrule
$\pi$& E2E AD Policy \\
$\mathcal{F}$& Case Difficulty Score\\
$\theta$& Generalist Parameters\\
$\Delta\theta$& Specialist Parameters\\
$\mathcal{D}_{\text{Train}}/\mathcal{D}_{\text{Hard}}/\mathcal{D}_{\text{RL}}$& Train / Hard / RL Data\\\midrule
$\mathcal{R}(\cdot)$& Reward Function \\
$\mathcal{C}(\cdot)$& Cost Function \\
$\mathbf{1}(\cdot)$& Indicator Function \\
$\texttt{IS}(\cdot)$& Importance Sampling Weight \\
\midrule
$\mathbf{U}_\text{Test}$& Generalist Uncertainty\\
$\hat{\theta}_{\text{Hard}}$& Estimated CDF Parameters\\
$p/\mathbf{P}$& PDF / CDF \\
\bottomrule
\end{tabular}
}
\vspace{-0.1cm}
\label{table:notation3}
\end{table}

\begin{table}[htp]
\caption{{Input/Output Notations}}
\centering
\resizebox{\linewidth}{!}{
\begin{tabular}{llll}
\toprule
\multirow{2}{*}{Notation}& \multirow{2}{*}{Meaning}            & \multicolumn{2}{c}{Tensor Shape}  \\
                        &                                      & nuPlan & CARLA       \\ \midrule
$\textbf{X}_{\text{Img}}$& Image Input & $8\times H \times W \times 3$&$H \times W \times 3$\\
$\textbf{X}_{\text{LiDAR}}$& LiDAR Input & -&$256\times 256\times n_{\text{pt}}$\\
$\textbf{X}_{\text{Ego}}$& Ego state Input &6 & 6\\\midrule
$\hat{\textbf{X}}_{\text{Agent}}$& Det./Track Output & \multicolumn{2}{c}{$N_A \times T_h \times d_A$} \\
$\hat{\textbf{X}}_{\text{Map}}$& Mapping Output & \multicolumn{2}{c}{$N_M \times L_M \times d_M$} \\
\textbf{Y},$\hat{\textbf{Y}}$& Planning Output & \multicolumn{2}{c}{$M\times T \times 3$} \\
\bottomrule
\end{tabular}
}
\label{table:notation2}
\vspace{-0.1cm}
\end{table}

\subsection{Analytical Results}\label{supp:theorem} 
\begin{theorem}[PAC-Bayes Generalization Bound for Adapter Ensembles]
\label{theorem1}
Let $\mathcal{H}$ be a hypothesis space of neural networks fine-tuned via LoRA, and let $\mathcal{D}$ be a data distribution. Let $P$ be a prior distribution over $\mathcal{H}$ (e.g., the base model), and let $Q$ be the posterior defined as the uniform distribution over an ensemble of $k$ LoRA-tuned models $h_1, \ldots, h_k$. 

Assume a bounded loss function $\ell: \mathcal{H} \times \mathcal{X} \times \mathcal{Y} \to [0, 1]$. Then, for any $\delta \in (0, 1)$, with probability at least $1 - \delta$ over a sample $S \sim \mathcal{D}^n$, the generalization loss of the ensemble satisfies:
\[
\mathbb{E}_{h \sim Q}[\mathcal{L}_{\mathcal{D}}(h)] \leq \mathbb{E}_{h \sim Q}[\mathcal{L}_S(h)] + \sqrt{ \frac{KL(Q \| P) + \log \frac{2\sqrt{n}}{\delta} }{2n} }
\]
\end{theorem}

\begin{proof}
This follows directly from the PAC-Bayes bound~\cite{langford2002pac}. The ensemble posterior $Q$ is defined as: $Q = \frac{1}{k} \sum_{j=1}^{k} \delta_{h_j},$ where $\delta_{h_j}$ is a Dirac distribution centered on model $h_j$. The empirical risk is:
$
\mathbb{E}_{h \sim Q}[\mathcal{L}_S(h)] = \frac{1}{k} \sum_{j=1}^{k} \mathcal{L}_S(h_j)
$
and $KL(Q \| P)$ is the KL-divergence between the posterior ensemble and the prior. If the LoRA-tuned models remain close to the base model, $KL(Q \| P)$ remains small.

Therefore, the bound quantifies the generalization error of the ensemble in terms of its average empirical risk and divergence from the prior, guaranteeing improved performance when individual models correct different failure cases via ensembling.
\end{proof}

\begin{theorem}[Pickands-Balkema-De Haan Theorem ~\cite{pickands1975statistical}]
\label{theorem2}
Let \( X \) be a real-valued random variable with cumulative distribution function \( F \), and let \( x_F = \sup \{ x : F(x) < 1 \} \) be the right endpoint of its support. Define the conditional excess distribution function over a threshold \( u \) as
\[
F_u(y) = P(X - u \leq y \mid X > u) = \frac{F(u + y) - F(u)}{1 - F(u)},
\]
where $y\ge0$. Then, for a large class of underlying distributions \( F \), there exists a positive scaling function \( \beta(u) > 0 \) and a shape parameter \( \xi \in \mathbb{R} \) such that
\[
\lim_{u \to x_F} \sup_{y \in [0, x_F - u)} \left| F_u(y) - G_{\xi, \beta(u)}(y) \right| = 0,
\]
where \( G_{\xi, \beta}(y) \) is the Generalized Pareto Distribution (GPD).
\end{theorem}

\begin{IEEEbiography}[{\includegraphics[width=1in,height=1.25in,clip,keepaspectratio]{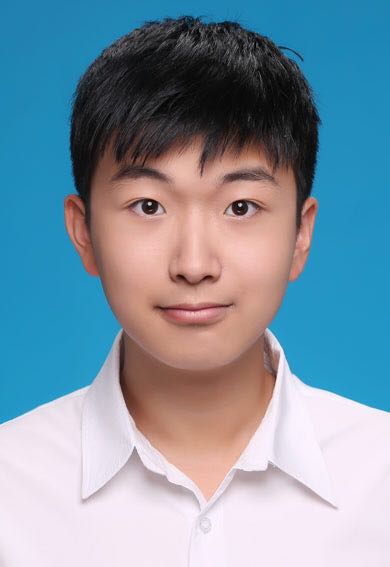}}]{Haochen Liu} received the B.E. degree from the School of Automation Science and Electrical Engineering, Beihang University, Beijing, China, in 2021. He is currently pursuing the Ph.D. degree with the School of Mechanical and Aerospace Engineering, Nanyang Technological University, Singapore. He won Frist Place
with Waymo Open Dataset Challenge 2024.
His current research interests include deep learning-enabled motion prediction and decision-making.
\end{IEEEbiography}
\vspace{-20pt}
\begin{IEEEbiography}[{\includegraphics[width=1in,height=1.25in,clip,keepaspectratio]{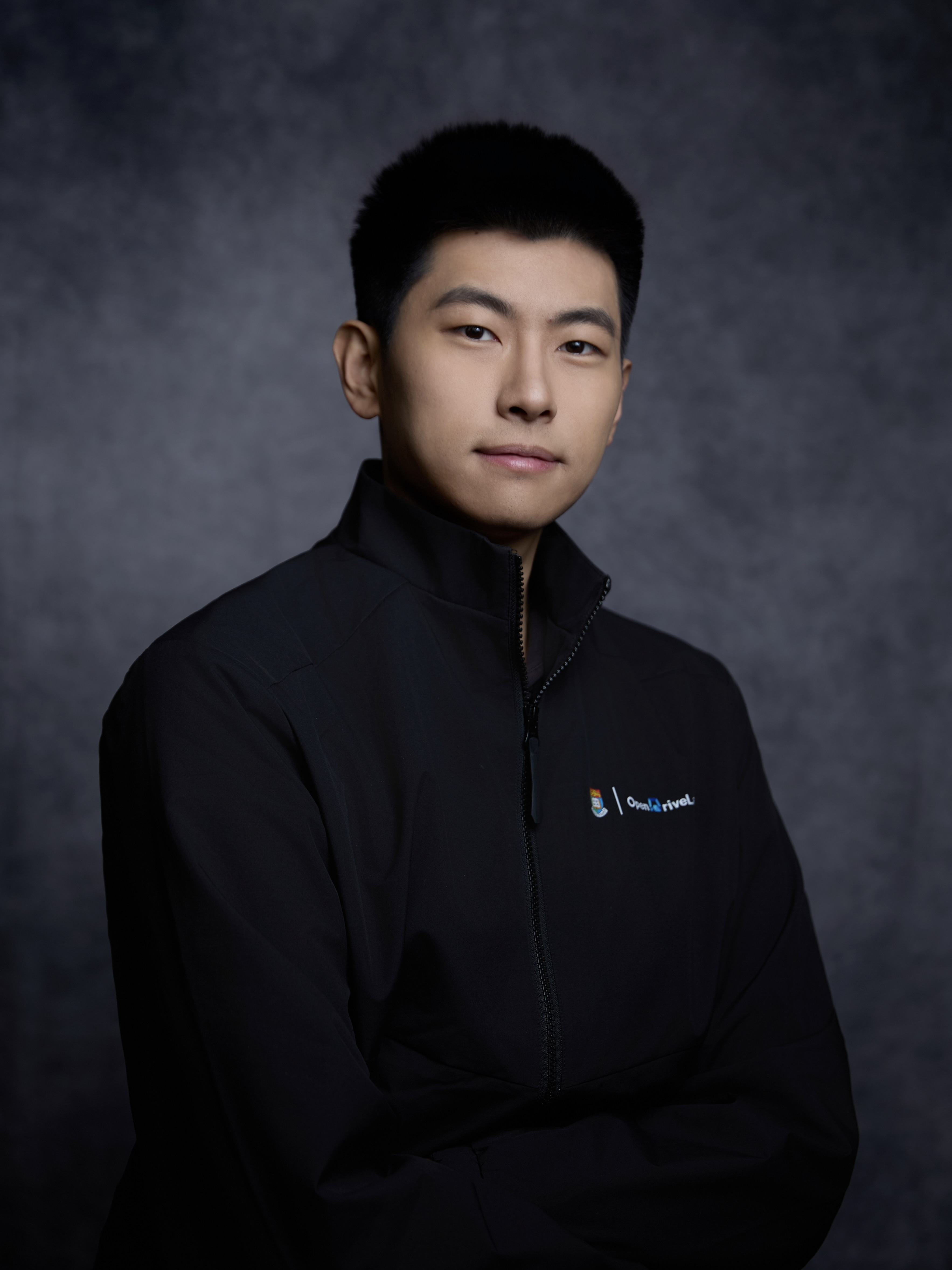}}]{Tianyu Li} is a Ph.D. student at Fudan University and the Shanghai Innovation Institute. He received his B.E. and M.S. degrees in Computer Science and Engineering from Beihang University, Beijing, China. He also serves as a researcher at OpenDriveLab, the University of Hong Kong. His work focuses on end-to-end autonomous driving.
\end{IEEEbiography}
\vspace{-20pt}
\begin{IEEEbiography}[{\includegraphics[width=1in,height=1.25in,clip,keepaspectratio]{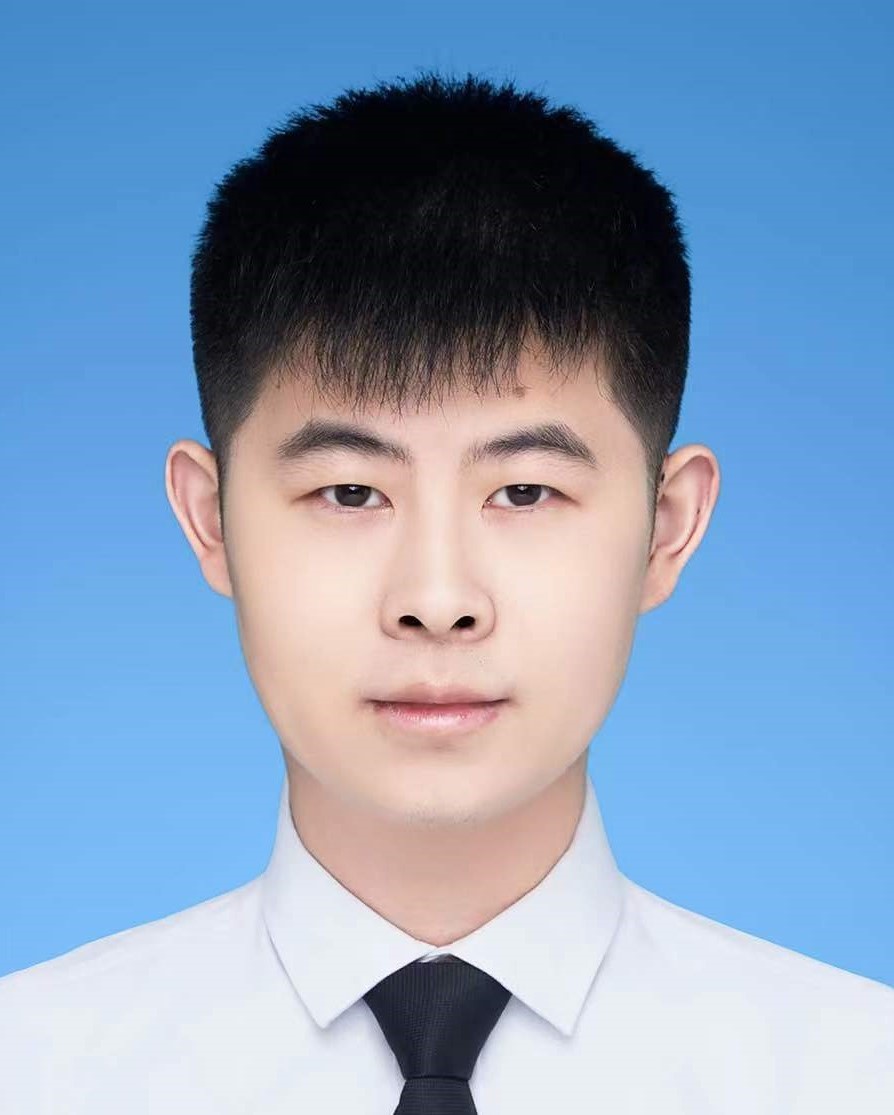}}]{Haohan Yang} received the B.S. degree in mechanical engineering from Dalian University of Technology, Dalian, China, in 2018, and the M.S. degree in vehicle engineering from Shanghai Jiao Tong University, Shanghai, China, in 2021. He is currently pursuing the Ph.D. degree with the School of Mechanical and Aerospace Engineering, Nanyang Technological University, Singapore. 
His current research interests include driver monitoring, human-in-the-loop AI, and continual learning.
\end{IEEEbiography}
\vspace{-20pt}
\begin{IEEEbiography}
[{\includegraphics[width=1in,height=1.25in,clip,keepaspectratio]{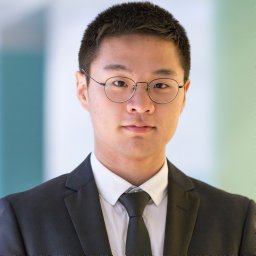}}]
{Li Chen} is currently a Ph.D. student at Department of Computer Science, the University of Hong Kong. He received the B.E. in mechanical engineering from Shanghai Jiao Tong University, and the M.S. in Robotics from the University of Michigan, Ann Arbor, USA. He also serves as a researcher with OpenDriveLab at Shanghai AI Laboratory. His research interests include autonomous driving and computer vision.
\end{IEEEbiography}
\vspace{-20pt}
\begin{IEEEbiography}[{\includegraphics[width=1in,height=1.25in,clip,keepaspectratio]{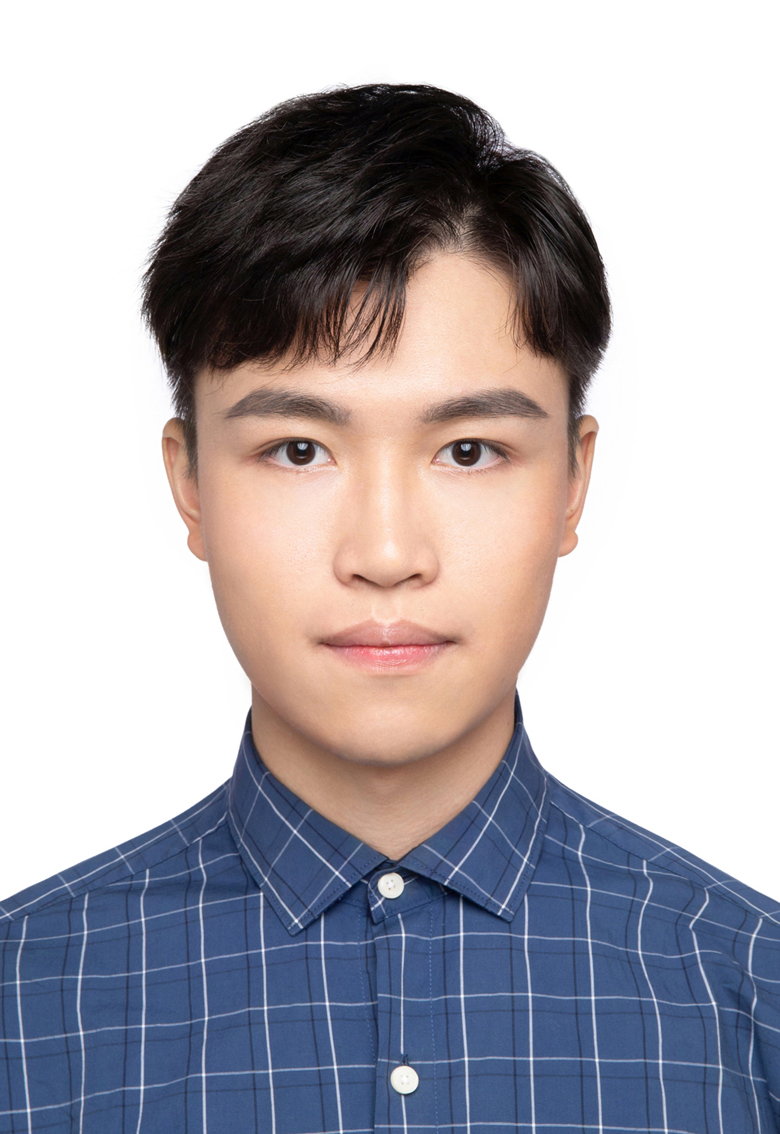}}]{Caojun Wang} received the B.S. degree (cum laude) from the School of Mechatronic Engineering And Automation, Shanghai University, Shanghai, China, in 2023. He is currently pursuing the Ph.D. degree in Tongji University and Shanghai Innovation Institute. His research interests include reinforcement learning and autonomous driving.
\end{IEEEbiography}
\vspace{-20pt}
\begin{IEEEbiography}[{\includegraphics[width=1in,height=1.25in,clip,keepaspectratio]{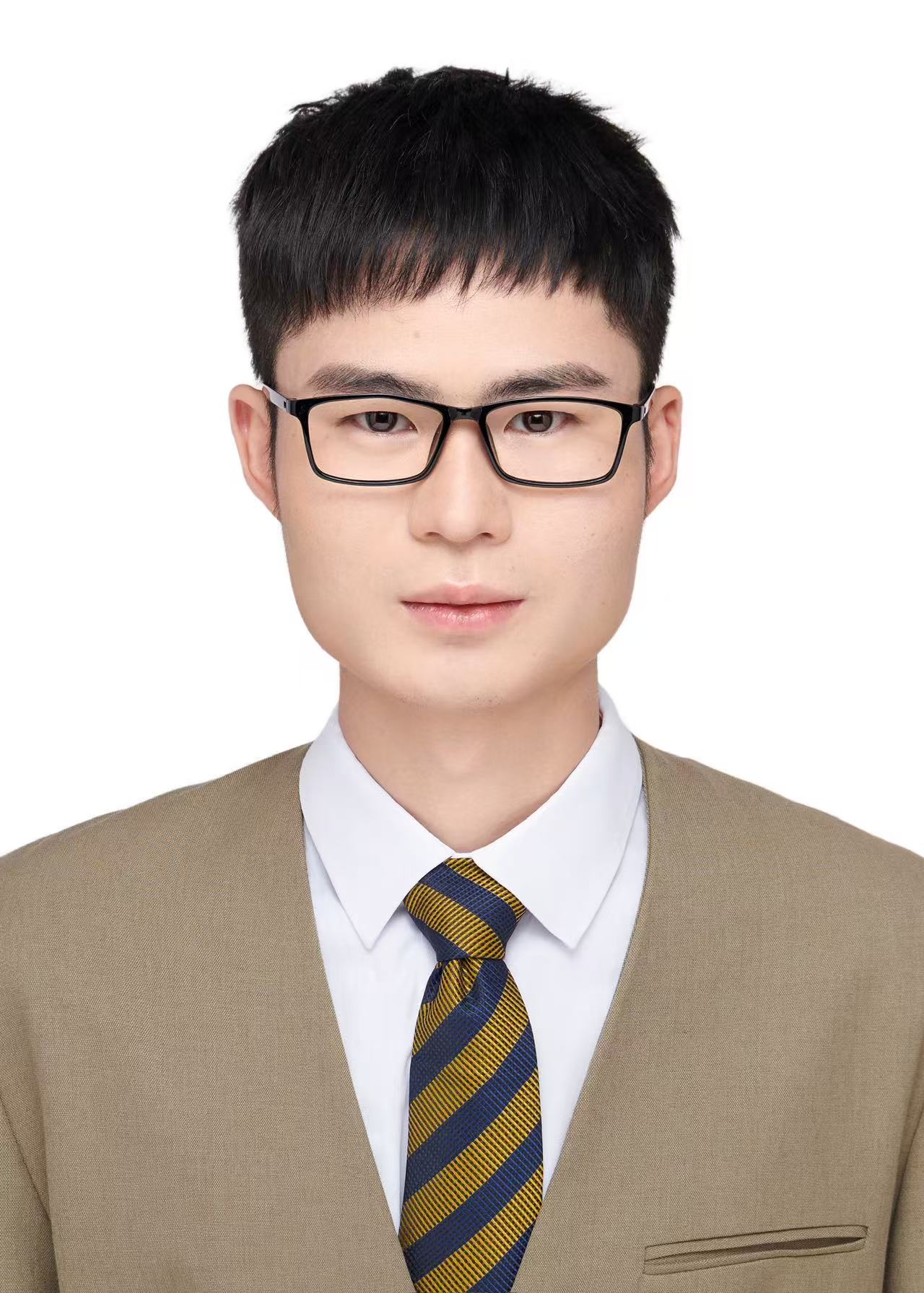}}]{Ke Guo} received the BE degree in automation from Zhejiang University, and the PhD degree in Computer Science from the University of Hong Kong. He is currently a research fellow in AutoMan Lab of Nanyang Technological University.  His research interests include autonomous driving and traffic simulation.
\end{IEEEbiography}
\vspace{-20pt}
\begin{IEEEbiography}[{\includegraphics[width=1in,height=1.25in,clip,keepaspectratio]{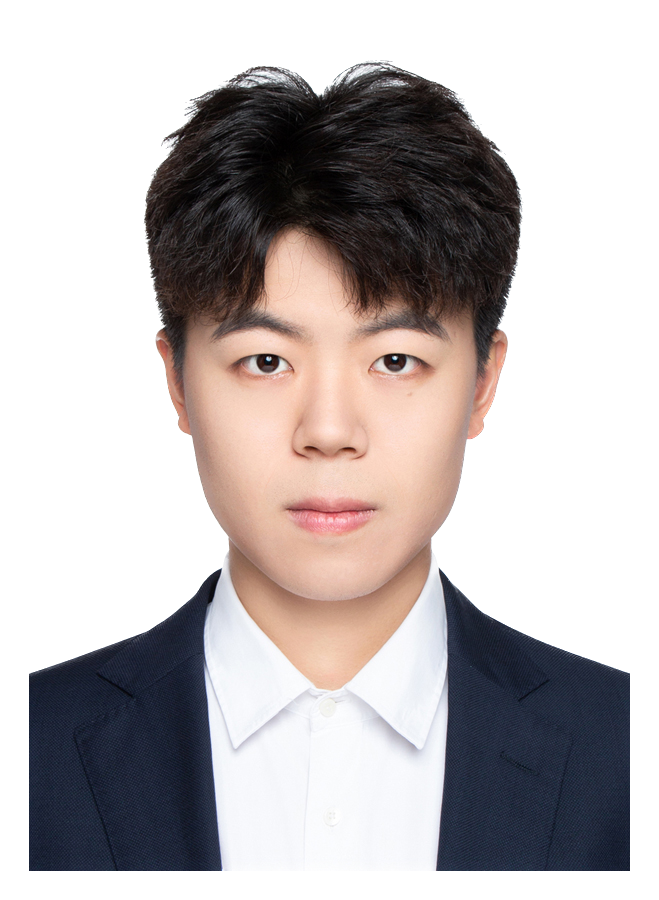}}]{Haochen Tian} received the BE degree in Spatial Informatics and Digitalized Technology from Wuhan University, Wuhan, China. He is
currently working toward the Ph.D. degree with National Laboratory of Multimodal Artificial Intelligence Systems (MAIS), Institute of Automation, Chinese Academy
of Sciences (CASIA), Beijing, China. Additionally, he serves
as an intern with OpenDriveLab. His current research interests include end-to-end autonomous driving and multimodal large language models.
\end{IEEEbiography}
\vspace{-20pt}
\begin{IEEEbiography}[{\includegraphics[width=1in,height=1.25in,clip,keepaspectratio]{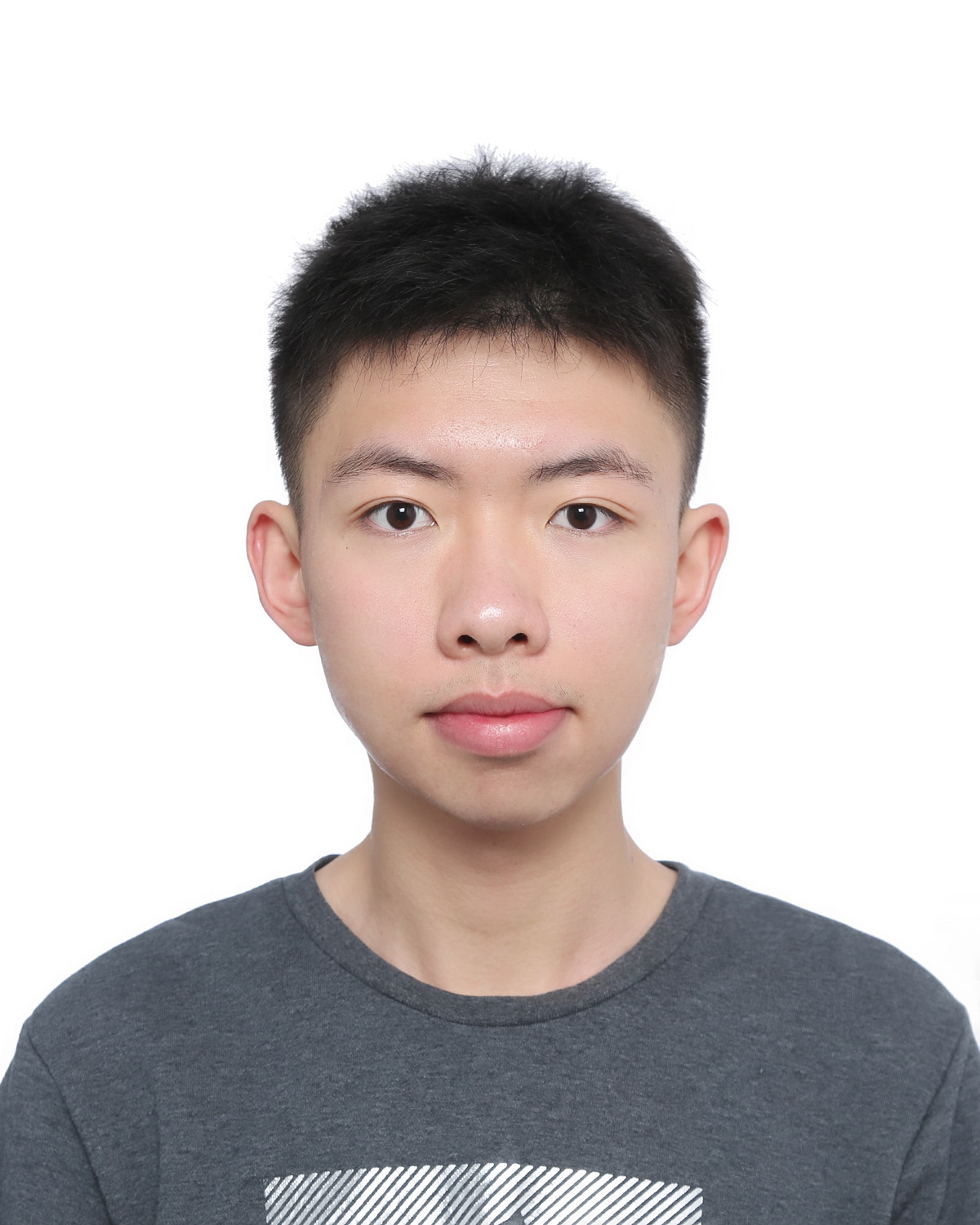}}]{Hongchen Li} received the B.S. degree Beijing Jiaotong University, Beijing, China in 2022. He is currently working toward the Ph.D. degree with Tongji University, Shanghai, China. His main research interests include end-to-end automated driving, deep reinforcement learning and motion prediction.
\end{IEEEbiography}
\vspace{-20pt}
\begin{IEEEbiography}
[{\includegraphics[width=1in,height=1.25in,clip,keepaspectratio]{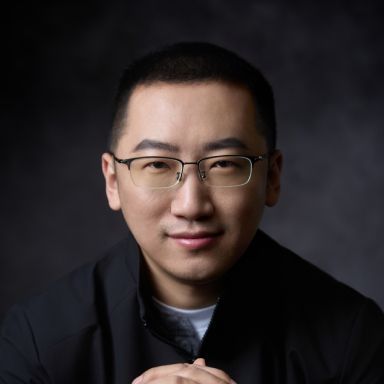}}]
{Hongyang Li} (S'13-M'20-SM'23) received Ph.D. in Computer Science from The Chinese University of Hong Kong in 2019. He is currently an Assistant Professor at Musketeers Foundation Institute of Data Science, University of Hong Kong and has led OpenDriveLab since 2021. His research focus is on autonomous driving and embodied AI. His expertise focuses on perception and cognition, end-to-end autonomous driving, and foundation models. He won as PI the CVPR 2023 Best Paper Award, and proposed BEVFormer that claimed Top AI 100 Papers 2022. He serves as Area Chair for NeurIPS, CVPR and referee for Nature Communications, Working Group Chair for IEEE Standards P3474. 
\end{IEEEbiography}
\vspace{-20pt}
\begin{IEEEbiography}[{\includegraphics[width=1in,height=1.25in,clip,keepaspectratio]{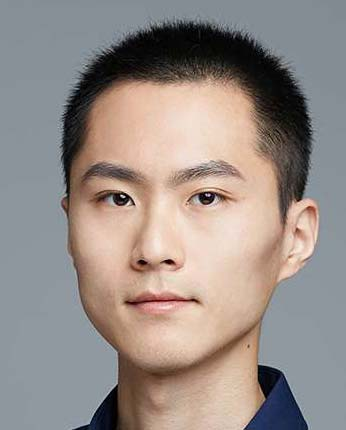}}]{Chen Lv} (Senior Member, IEEE) received a Ph.D. degree from the Department of Automotive Engineering, Tsinghua University, China, in 2016. 
	
He is currently an Associate Professor at the School of Mechanical and Aerospace Engineering, Nanyang Technology University, Singapore. He also holds joint appointments as the Cluster Director in Future Mobility Solutions at ERI@N, Thrust Lead in Smart Mobility and Delivery, Continental-NTU Corp Lab, and the Program Lead in Next Generation AMR, Schaeffler-NTU Joint Lab. His research focuses on advanced vehicles and human-machine systems, where he has contributed over 200 papers and obtained 12 granted patents in China. Dr. Lv serves as an Associate Editor for \emph{IEEE Transactions on Intelligent Transportation Systems}, \emph{IEEE Transactions on Vehicular Technology}, \emph{IEEE Transactions on Intelligent Vehicles}, etc.
\end{IEEEbiography}

\end{document}